\documentclass[review]{elsarticle}
\pdfoutput=1

\journal{Aerospace Science and Technology}

\usepackage{amsmath}

\usepackage{amsmath}

\usepackage{graphicx}
\usepackage{subfigure}
\usepackage{empheq}
\usepackage{booktabs}
\usepackage{nomencl}
\makenomenclature
\usepackage{framed}
\usepackage{longtable}
\usepackage{array}
\usepackage{amssymb}
\usepackage{color}
\usepackage{multirow}
\usepackage{setspace}%使用间距宏包
\makenomenclature
\usepackage{tabularx}
\usepackage{changepage}
\usepackage{mathdots}
\usepackage{makecell}

\begin{document}

\begin{frontmatter}
%\title{Augmented Imagefication:\\ A Data-driven Fault Detection Method for Aircraft Air Data Sensors\tnoteref{mytitlenote}}
\title{Fault Detection and Classification of Aerospace Sensors using a VGG16-based Deep Neural Network\tnoteref{mytitlenote}}

\tnotetext[mytitlenote]{This work was sponsored by Shanghai Sailing Program under Grant No. 20YF1402500, and Shanghai Natural Science Fund under Grant No. 22ZR1404500.}

\author[fd]{Zhongzhi Li}
\ead{sau_lzz@foxmail.com}

\author[tj]{Yunmei Zhao}
\ead{yunmeizhao@tongji.edu.cn}

\author[fd]{Jinyi Ma}
\ead{jyma21@m.fudan.edu.cn}

\author[fd]{Jianliang Ai}
\ead{aijl@fudan.edu.cn}

\author[fd]{Yiqun Dong\corref{mycorrespondingauthor}}
\cortext[mycorrespondingauthor]{Corresponding author.}
\ead{yiqundong@fudan.edu.cn}

\address[fd]{Department of Aeronautics and Astronautics, Fudan University, Shanghai 200433, China}
\address[tj]{School of Aerospace Engineering and Applied Mechanics, Tongji University, Shanghai 200092, China}

\begin{abstract}
Compared with traditional model-based fault detection and classification (FDC) methods, deep neural networks (DNN) prove to be effective for the aerospace sensors FDC problems. However, time being consumed in training the DNN is excessive, and explainability analysis for the FDC neural network is still underwhelming. A concept known as imagefication-based intelligent FDC has been studied in recent years. This concept advocates to stack the sensors measurement data into an image format, the sensors FDC issue is then transformed to abnormal regions detection problem on the stacked image, which may well borrow the recent advances in the machine vision vision realm. Although promising results have been claimed in the imagefication-based intelligent FDC researches, due to the low size of the stacked image, small convolutional kernels and shallow DNN layers were used, which hinders the FDC performance. In this paper, we first propose a data augmentation method which inflates the stacked image to a larger size (correspondent to the VGG16 net developed in the machine vision realm). The FDC neural network is then trained via fine-tuning the VGG16 directly. To truncate and compress the FDC net size (hence its running time), we perform model pruning on the fine-tuned net. Class activation mapping (CAM) method is also adopted for explainability analysis of the FDC net to verify its internal operations. Via data augmentation, fine-tuning from VGG16, and model pruning, the FDC net developed in this paper claims an FDC accuracy 98.90\% across 4 aircraft at 5 flight conditions (running time 26 ms). The CAM results also verify the FDC net w.r.t. its internal operations.
\end{abstract}

\end{frontmatter}

\section{Introduction}
\subsection{Motivation}

Aerospace sensors provide measurements for the aircraft states which include airspeed, attitude angles, load factors, etc. It is found that many catastrophic flight accidents were caused by faults in the aerospace sensors measurements \cite{X31_crash, AF330, B737Max}. It is imminent to develop fault detection and classification (FDC) techniques for the aerospace sensors. 

Hardware redundancy (HR) has been widely used in commercial airlines for the FDC problem. HR consists of installing multiple sets of sensors on the aircraft that provide redundant measurements of the states. A voting logic is designed to monitor the outputs from all sensors, which isolates the sensor(s) with fault, and decides the correct state value using remaining sensor(s) \cite{HR-1,HR-2,HR-3}. However, one issue with the HR-based FDC scheme is the cost and weight penalty (due to redundant sensors installations). Recent accidents caused by aerospace sensors in commercial airlines \cite{AF330, B737Max} also indicate that HR-based strategy is still insufficient in addressing the FDC problems. 

Alternative to HR, analytical redundancy (AR) is also widely used in FDC researches \cite{MB-review}. AR typically adopts model-based approaches, wherein the model depicts the evolution of flight states, sensors characteristics, etc. For a certain sensor, the associated state is firstly estimated using the measurements of other sensors. The estimation is then compared with the sensor's output to generate a residual. If the residual exceeds a predefined bound, fault is defined to be detected for that sensor. However, model-based AR hinges on the model of flight states/sensors characteristics, which are difficult to accurately attain. 

Another line of the AR-based FDC algorithms adopts the recent advances in artificial intelligence, and particularly deep neural networks (DNN) \cite{own_IEEE_TAES}. In particular, the FDC problem is modeled as a mapping, of which the inputs are available data from sensors, and output the FDC labels. One significant problem of DNN-based FDC, however, is the ``black box" issue; i.e., it is difficult to develop a universal rule in tuning the DNN architectures, and mathematical operations enclosed within the DNN are unclear. DNN-based FDC techniques, therefore, has not been widely adopted (despite its high performance). 

In other related fields (e.g. machine vision), explainability analysis of DNN has been investigated. In particular, ablation studies are used to verify the architectural design (e.g. number of layers/nodes) of the DNN \cite{ablationstudy}, and class activation mapping (CAM) has been used in analyzing the convolutional operations enclosed within the DNN \cite{zhou2016}. Fine-tuning/Transfer learning is also widely used to tune the DNN architectures/parameters with much less time being consumed \cite{transferlearning}. Recent advances in both academic and industrial fields prove the efficacy of explainability analysis and fine-tuning/transfer learning methods for DNN studies in the machine vision realm.     

Following the DNN advances in the machine vision realm, one question naturally arises as can the mature methods (e.g. CAM, fine-tuning) be implemented in developing and analyzing the DNN for FDC problems. For such purpose, imagefication-based intelligent FDC concept has been proposed. This concept advocates to stack the sequential measurement data of different aerospace sensors into an image format (i.e. a `sensors data image', SDI). The sensors FDC is then transformed to abnormal region(s) detection on the stacked image, which may well borrow the recent DNN advances in the machine vision realm. Recent research endeavors along this line yield promising results for both FDC performances as well as the DNN explainability analysis \cite{AST, EAAI, IJAE-yiming, dong2021deep, IJAE-robust}. 

Development (And analysis) of DNN in the imagefication-based intelligent FDC scheme, nevertheless, depends on the pixel size of the stacked SDI. Although current researches have considered all sensors (12 states and 3 load factors), the SDI stacking is separated for each group of the states  (e.g. individually for air data, attitude angles), rendering the SDI size low (3$\times$31) and distorted as compared with the image size in machine vision realm. The FDC neural network must be designed and trained from scratch, which is time consuming. Moreover, the convolutional kernel size has to maintain small ($2\times2$) for a ``valid" scanning within the SDI image; this may render the convolutional features extraction inefficient, and depth of the DNN layers insufficient. Although recent advances in imagefication-based intelligent FDC claims a fair 90\% accuracy for the air data sensors FDC problem, further improvements were found difficult \cite{dong2021deep, IJAE-robust}. 

We thus summarize the motivation of this work: As the current dilemma of imagefication-based FDC research originates from the low SDI size, we hope to augment the SDI to a relatively larger size, so the DNN performance may arise as larger convolutional kernels and deeper DNN architecture can be used. Also, preferably the augmented SDI size should match the input of typical pre-trained DNN architectures in the machine vision realm. The FDC neural network then may borrow the fine-tuning/transfer learning strategies from the machine vision realm, which may significantly reduce the time being consumed in training the FDC neural network. Following these motivations, we introduce the most recent advances for the imagefication-based intelligent FDC research in this paper. 

\subsection{Related Works}
\subsubsection{Model-based and Data-driven FDC Studies}

A plethora of model-based FDC methods were found to use Kalman filtering (KF), e.g. the extended KF \cite{KF-27, KF-18, KF-AST}, the unscented KF \cite{KF-11}, and the theoretical analysis of an adaptive three-step KF \cite{KF-7}. Other model-based FDC methods adopted robust synthesis theory in \cite{Hinf, Hinf-1, Hinf-2, AST-robust}, wherein $H\infty$-based filters were designed to output the fault directly. In \cite{MHE-1, MHE-2, MHE-3}, moving horizon estimator was developed, which compensated for both sensors faults and wind speed estimation in the fault tolerant control. A scheme designed particularly for systems with time-scale dynamics (e.g. phugoid and short periods in the aircraft's longitudinal plane) was discussed in \cite{NGA-1, NGA-2}, where in both nonlinear geometric approach and singular perturbation techniques were involved. Other works adopted barrier function-based learning observers \cite{AST-barrier} and set-value observers \cite{SVO} in the work, which was claimed to significantly decrease the FDC false alarm rates. However, model-based FDC methods typically involves a model that is derived from the aircraft dynamics/sensors characteristics. The model (and the coefficients enclosed within) are difficult to accurately attain. \textit{Ad hoc} tuning is widely needed in adjusting the parameters adopted in the model-based FDC methods, wherein a universal rule was rarely found (should it exist at all). This therefore hinders the development and applications of model-based FDC methods. 

Another recent and more dynamic line of the FDC development adopts data-driven, and mostly neural networks. In \cite{NN-26, NN-29} fully connected cascaded neural network was adopted, wherein FDC problems for aircraft inertial measurement unit (IMU) were discussed. Similar problems were also studied in \cite{NN-17, NN-19, MJ-4} using feed-forward neural network. In \cite{NN-24, NN-31} the sensors measurement residual was generated using neural network-based observers, and parameters of the neural network was adjusted online using KF \cite{NN-24}. Neural networks were also used to establish nonlinear identification models in \cite{DataDriven-1, DataDriven-3, DataDriven-4}, which being further used as a state observer to generate the residual. Recent neural networks development advocates the DNN technique in many academic/industrial fields \cite{own_IEEE_TAES}. DNN typically has deeper architecture which are activated using designated functions (e.g. ReLU \cite{deeplearningbook}). More dedicated features extraction operations have also been developed which includes convolution and long-short time memory (LSTM) blocks \cite{deeplearningbook}. Early DNN-based FDC methods were found in \cite{RNN-1, RNN-2, RNN-3}, wherein recurrent neural networks were used. Later developments adopt both LSTM and convolution operations in designing the FDC neural network \cite{MJ-3, MJ-7}. However, despite the rapid developments of DNN-based FDC method, the weakness of this method is also overwhelming: It is rather time-consuming to train the DNN architectures; research efforts pertaining to explainability analysis of the FDC neural network are also rare, which renders the DNN-based FDC techniques less convincing (despite its high performance). 

\subsubsection{Representative DNN Studies in the Machine Vision Realm}

The dilemma of data-driven FDC researches originates in the DNN architectural design/analysis, i.e time-consuming DNN training, cubersome DNN architecture size, and unclear operations enclosed within the DNN. Recent advances in the machine vision realm, however, has claimed promising results for these subjects. 

\textbf{Fine-tuning/Transfer learning:} The DNN training in machine vision may be much more time-consuming when compared with the aerospace sensors FDC researches, as much larger size (e.g. 1080p, 4K) and data volume (to fully learn to extract the features) need to be considered. Similar features extracted from the input images, however, are considered relevant and useful in different tasks (e.g. image classification and object detection). Many researchers thus advocate to absorb DNN that has been pre-trained from a large dataset, and adapt it to different tasks. In this process, the architecture of the DNN may remain identical as the original one (but internal parameters including biases, weights are re-trained), or the less-relevant structures are cropped and the training efforts are imposed on the remaining core structures only. The former method is defined as fine-tuning, and the latter transfer learning, both drastically decreases the time needed in training the DNN. Many fine-tuning/transfer learning works in the machine vision realm adopt VGG16 \cite{simonyan2014very}, which was pre-trained using ImageNet \cite{deng2009imagenet}, a dataset that consists of more than 14 million images. VGG16 won the championship in the 2014 ILSVRC challenge with a claimed image classification accuracy at 92.3\% (top 5) for 1,000 image categories. VGG16 has laid the basis as an effective and universal features extractor in the development (e.g. fine-tuning, transfer learning) of many machine vision DNN such as the Yolo series \cite{YOLO-1, YOLO-2, YOLO-3}, the Mask-CNN series \cite{fast-rcnn, faster-rcnn, mask-rcnn} etc. 

\textbf{Model pruning/compression:} Although deeper architecture is usually preferred (and more effective) in DNN, previous research indicates that effective operations (e.g. convolutional filters) enclosed within the net may be only 5\% \cite{denil2013predicting}. Model pruning is thus widely used in truncating redundant operations from the DNN. Model pruning consists of 3 stages: pre-training, pruning, and fine-tuning. In the pre-training stage, a model needs to be constructed, e.g. by absorbing the existing pre-trained net \cite{luo2017thinet}, or training a neural network from scratch \cite{liu2017learning}. In the pruning stage, filters (e.g. convolutional kernels) evaluation models are designed to remove the less-relevant operations; previous researches adopt the $L_1$ norm \cite{li2016pruning} or the size of the network normalization layer scaling factor \cite{liu2017learning}. Finally, to guarantee performance of the DNN, the net is usually fine-tuned using the dataset (same data as in the pre-training stage). In \cite{luo2017thinet}, it was found that the proposed pruning method achieves 16.63 times size compression on VGG16, with only 0.52\% top-5 accuracy drop.

\textbf{Explainability analysis:} Convolutional neural networks (CNN) are widely used in machine vision for image classification, object detection, etc. CNN hinges on the convolutional operations to extract cascaded features from the input image. Many research efforts have been focused on explaining the convolutional operations, i.e. visualizing the features that are reported in the convolutional kernels. In \cite{JAIS} features visualization was adopted to explain the operation of a certain convolutional node enclosed in the CNN layers. Later works proposed class activation mapping (CAM) \cite{zhou2016} and its variants \cite{selvaraju2017, chattopadhay2018, omeiza2019, wang2020, wang2020ss, ramaswamy2020} to synthesize all the convolutional kernels effects in a certain CNN layer. Typically, CAM reveals the region of attention (ROA) of a certain CNN layer on the input image. If the ROA corresponds to the regions of (humans') interest for a certain machine vision task (e.g. to decide if an object is a cat/dog based on the nose/mouth appears on the image), CAM results are then claimed to be convincing, which in return verifies the performance of the CNN--it is indeed working (extracting the designated features) as humans intend to. CAM analysis has been considered as a must in many machine vision researches, which greatly solidifies the associated DNN architectures (hence the DNN performances). 

\subsubsection{Development of the Imagefication-based Intelligent FDC}

In referring to the DNN advances in the machine vision researches (e.g. fine-tuning to reduce the time being consumed in DNN training, and CAM to verify the DNN performance), a concept known as imagefication-based intelligent FDC has been studied in \cite{AST, EAAI, IJAE-yiming, dong2021deep, IJAE-robust}. In \cite{AST}, the authors proposed SDI, which down-samples the sequential sensors measurement data within a certain history time window, and stack the data into an image format. Aircraft icing detection and classification problem was studied in comparison with a state-of-the-art $H\infty$-based algorithm, wherein the imagefication-based intelligent FDC approach shows more promising results. In \cite{EAAI} SDI was furtherly used for the detection and classification of sensors faults, actuators failures, and icing, which claimed 80\% accuracy for a total of 13 fault cases. In \cite{IJAE-yiming} the FDC problem for IMU failures was studied, and in \cite{dong2021deep, IJAE-robust} in answering to the Boeing B737MAX flight accidents, researches particularly for the aircraft air data sensors was studied, which claimed an accuracy at 90\% across 5 aircraft at 6 diverse flight conditions. Features visualization and CAM analysis was also adopted in \cite{IJAE-robust} to explain the convolutional operations enclosed within the proposed FDC neural network, which yields promising results. 

Nevertheless, previous works along the imagefication-based intelligent FDC research line advocate to stack the SDI separately for different groups of the aircraft states. In \cite{dong2021deep, IJAE-robust}, the air data sensors FDC problem was studied, SDI was stacked for 3 air data states which include airspeed, angle of attack, and sideslip angle; a 30s time window and 1Hz was also used to down-sample the sensors data; the SDI thus claimed a $3\times31$ pixel size, which is relatively low and uncommonly distorted as compared with regular images. All the FDC neural networks in \cite{AST, EAAI, IJAE-yiming, dong2021deep, IJAE-robust} therefore needs to be designed and trained from scratch, which is time-consuming. Moreover, as the SDI pixel size was relatively low and valid convolutional kernel scanning was adopted in the FDC neural network, small convolutional kernel size ($2\times2$) was adopted, which may render the convolutional operations enclosed within the FDC neural network less efficient. Depth of the FDC neural network also has to remain shallow as the feature map size exported in the convolutional layers (and imported as input for the next layer) drops rapidly due the valid convolutional kernel scanning. All these factors hinder the performance of the imagefication-based intelligent FDC method. In the latest research \cite{IJAE-robust}, although the FDC accuracy was found fair at 90\%, any further improvements, however, was very difficult. 

\subsection{Overview of This Paper}

\begin{figure}[t]
\centering
%trim={10pt 0pt 20pt 15pt},
\includegraphics[clip,width = 1
\columnwidth]{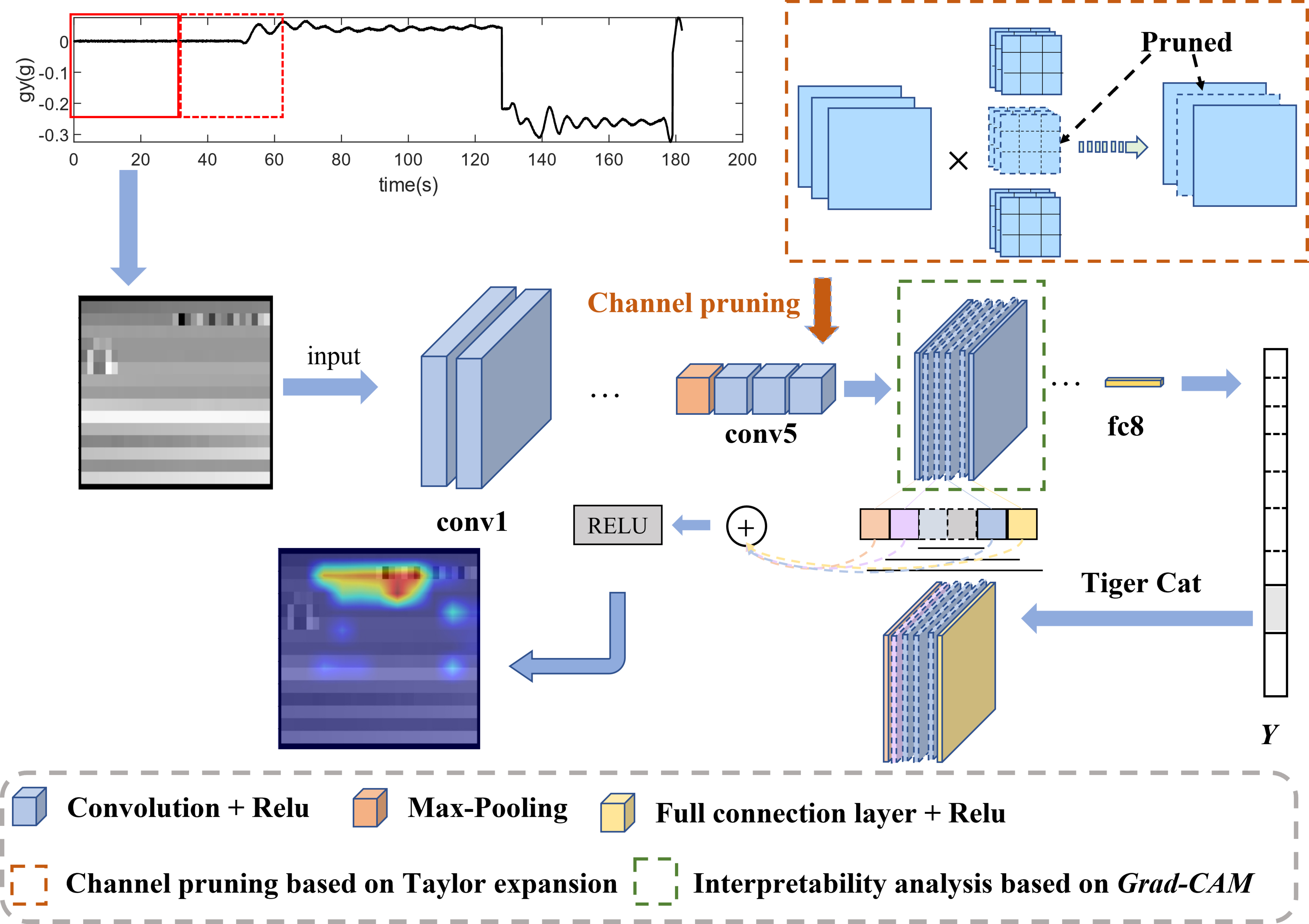}
\caption{Illustrations of the work presented in this paper: data augmentation, finetuning and model pruning, and CAM analysis.}
\label{figure_1}
\end{figure}

In response to the current dilemma of the imagefication-based intelligent FDC researches, we report our most recent research advances in this paper. A schematic plot of the work presented herein is shown in Figure \ref{figure_1}. We also summarize the highlights of this paper as threefold:
\begin{itemize}
\item \textbf{A data augmentation method}: We propose a method which effectively inflates the stacked SDI (size $15\times31$) to $224\times224$ pixel size, and larger convolutional kernel and deeper neural network architecture can be used; this size also matches the input of VGG16, the features-extractor net widely adopted in the machine vision realm. 
\item \textbf{VGG16-based FDC neural network}: The sensors FDC is transformed to an abnormal regions detection problem on the image; as the latter is a typical topic in the machine vision realm, we train the FDC neural network via fine-tuing the VGG16 from machine vision realm directly; this expedites the FDC net development and claims promising results; we also perform model pruning upon the fine-tuned net, to truncate its size to 5.60MB, compress the running time to 26ms, and slightly evaluate the testing accuracy to 98.90\% across 4 diverse aircraft at 5 flight conditions.   
\item \textbf{Explainability analysis and extensive tests}: We perform CAM analysis for the proposed (fine-tuned and pruned) FDC net, which verifies the convolutional features extraction enclosed within different layers of the net; we also perform extensive tests for the net using simulation and real flight data, both yields promising results.   
\end{itemize}

The rest of this paper is organized as follows. Section \ref{Section_II} defines the problem. Section \ref{Section_III} illustrates the database. The VGG16-based FDC network is developed in Section \ref{Section_IV}, wherein both data augmentation, network fine-tuning, and model pruning are introduced. In Section \ref{Section_V} we perform CAM analysis and extensive tests for the proposed FDC net. And finally in Section \ref{Section_VI}, conclusions of this paper are presented. 

\section{Problem Definition}\label{Section_II}

We start with air data evolution equations in defining the aerospace sensors FDC problem:
\begin{equation}
\begin{cases}
\begin{aligned}
    \dot{V}      = &(G_x-gS_{\theta})C_{\alpha}C_{\beta}+(G_y+gS_{\phi}C_{\theta)}S_{\beta} 
                    +(G_z+gC_{\phi}C_{\theta})S_{\alpha}C_{\beta}\\
    \dot{\alpha} = &(-G_xS_{\alpha}+G_zC_{\alpha}+gC_{\phi}C_{\theta}C_{\alpha}
                    +gS_{\theta}S_{\alpha})/VC_{\beta}+w_y\\&-(w_xC_{\alpha}+w_zS_{\alpha})S_{\beta}/C_{\beta}\\
    \dot{\beta}  = &[-(G_x-gS_{\theta})C_{\alpha}S_{\beta}+(G_y+gS_{\phi}C_{\theta})C_{\beta}
                    -(G_z+gC_{\phi}C_{\theta})S_{\alpha}S_{\beta}]/V\\&+w_xS_{\alpha}-w_zC_{\alpha}\\
\end{aligned}
\end{cases}
\label{translational_dynamics}
\end{equation}
where $S_*$ and $C_*$ denote \textit{sin} and \textit{cos} operations, respectively. In (\ref{translational_dynamics}), $\{V, \alpha, \beta\}$ denote airspeed, angle of attack, and sideslip angle; $g$ is the gravitational acceleration; $\{w_x, w_y, w_z\}$, $\{\psi, \theta, \phi\}$, and $\{G_x, G_y, G_z\}$ are rotational speeds, rotational angles, and load factors defined in the aircraft body axis. 

In (\ref{translational_dynamics}), rotational angles and speeds of the aircraft are coupled as:
\begin{equation}
\begin{cases}
\begin{aligned}
    \dot{\psi}   &= w_yS_{\phi}/C_{\theta} + w_zC_{\phi}/C_{\theta} \\
    \dot{\theta} &= w_yC_{\phi}-w_zS_{\phi}\\
    \dot{\phi}   &= w_x+w_yS_{\phi}S_{\theta}/C_{\theta}+w_zC_{\phi}S_{\theta}/C_{\theta}\\
\end{aligned}
\end{cases}
\label{rotational_kinematics}
\end{equation}
And aircraft motion equations are written as:
\begin{equation}
\begin{cases}
\begin{aligned}
    \dot{x} &= uC_{\theta}C_{\psi}+v(S_{\theta}S_{\phi}C_{\psi}-C_{\phi}S_{\psi})+w(S_{\theta}C_{\phi}C_{\psi}+S_{\phi}S_{\psi}) \\
    \dot{y} &= uC_{\theta}S_{\psi}+v(S_{\theta}S_{\phi}S_{\psi}+C_{\phi}C_{\psi})+w(S_{\theta}C_{\phi}S_{\psi}-S_{\phi}C_{\psi})\\
    \dot{z} &= -uS_{\theta}+vS_{\phi}C_{\theta}+wC_{\phi}C_{\theta}\\
\end{aligned}
\end{cases}
\label{XYZ}
\end{equation}
wherein the aircraft speed expressed in the body axes $\{u, v, w\}$ are:
\begin{equation}
\begin{cases}
\begin{aligned}
    u &= VC_{\alpha}C_{\beta} \\
    v &= VS_{\beta} \\
    w &= VS_{\alpha}C_{\beta} \\
\end{aligned}
\end{cases}
\label{VAB}
\end{equation}

Different sensors are installed onboard the aircraft to provide measurements of aircraft flight states. Typically, air data sensors (ADS) decide $\{V, \alpha, \beta\}$; inertial measurement unit (IMU) outputs $\{w_x, w_y, w_z\}$, $\{\psi, \theta, \phi\}$, and $\{G_x, G_y, G_z\}$; and position of the aircraft $\{x, y, z\}$ are reported via GPS. We consider FDC for ADS and IMU. The FDC problem is then defined as: To detect and classify the potential faults of ADS and IMU sensors, given real-time measurements of all the flight states as well as load factors. 

Whilst other works adopt model-based approaches in analyzing the dynamics/kinematics in Equations (\ref{translational_dynamics}$\sim$\ref{VAB}), we model the FDC problem as a mapping process, wherein DNN are used to implicitly extract and investigate the correlations of the sensors measurement data, and detect/classify the potential sensors fault. Particularly following the imagefication-based intelligent FDC research line, the sensors measurement data is stacked into an image (SDI). The problem we aim to address in this paper is then defined as: How to properly resize the SDI, and how to develop (and verify) the FDC neural network via the mature results/methods adopted in machine vision researches.

\section{Data Preparation}\label{Section_III}
\subsection{Diverse Aircraft and Flight Conditions}
Data is crucial for DNN training and validation. Most previous works discussed 1 aircraft only. We allocate both simulation and real flight data from 4 different aircraft which include 1 large cargo airplane (Y \cite{dong2016full}), 2 passenger aircraft (B$_1$ and B$_2$, \cite{hohndorf2017reconstruction, nelson1998flight}), and 1 general aviation (D \cite{AST, EAAI}). We also involve 5 different flight conditions to cover the aircraft's entire envelope, i.e. high and low altitudes for both cruise, manual free flight, and low-altitude landing/take-off cycle (LTO). Different control forms from both human pilot (manual) and automated control laws (auto-pilot, AP) are also considered. The details are in the Table \ref{aircraft_and_flightcondition}. 

\begin{table*}[h]
    \centering
    \caption{Different aircraft and flight conditions adopted in this paper.}
    \label{aircraft_and_flightcondition}
    \scalebox{0.637}{
    \begin{tabular}{c|lcccl}
    \hline
    \hline
    \textbf{Aircraft}   &\textbf{General configuration}   &\textbf{Weight}  &\textbf{Span} &\textbf{Data source} &\textbf{Flight condition; duration}\\
    \hline
    %\xrowht[()]{3.5pt}
    Y   
    &\multicolumn{1}{l}{\begin{tabular}[l]{@{}l@{}} large cargo airplane\\ 4 piston engines, high wing\end{tabular}}
    &41.0t   &38.0m   &simulation   &$\bullet$ low altitude, LTO, manual; 295min    \\  \hline  %\xrowht[()]{8pt}
    B$_1$  
        &\multicolumn{1}{l}{\begin{tabular}[l]{@{}l@{}} large passenger aircraft\\ 4 turbo engines, low wing\end{tabular}}
    &174t &59.6m   &simulation   
            &\multicolumn{1}{l}{\begin{tabular}[l]{@{}l@{}}$\bullet$ high altitude, cruise, AP; 151min\\ $\bullet$ low altitude, free flight, manual; 327min\end{tabular}} \\  \hline  %\xrowht[()]{3.5pt}
    B$_2$  
    &\multicolumn{1}{l}{\begin{tabular}[l]{@{}l@{}} large passenger aircraft\\ 2 turbo engines, low wing\end{tabular}}
     &44.6t &35.8m   &real flight  &$\bullet$ low altitude, LTO, manual; 67min                   \\  \hline  %\xrowht[()]{3.5pt}
    D   
    &\multicolumn{1}{l}{\begin{tabular}[l]{@{}l@{}} general aviation\\ 2 piston engines, high wing\end{tabular}}
    
    &3.12t   &19.8m   &simulation   &$\bullet$ high altitude, cruise, AP; 162min                      \\  \hline  %\xrowht[()]{3.5pt}
 \hline
    \end{tabular}
    }
\end{table*}

\subsection{Measurement Noises and Disturbances}
Both simulation and real flight data are considered in the paper. While measurement noises and disturbances exist naturally in the real flight, we adopt the model following \cite{MIL1797} in simulation. Dryden atmospheric disturbances are injected to perturb the flight states, on which the measurement noises are added to generate the noise-corrupted data. Measurement noises are assumed to follow Gaussian distribution. Standard deviations for the noise of each sensor are characterized in Table \ref{measurement_noise_std} \cite{KF-27}. 

\begin{table}[h]
    \centering
    \caption{Sensors noise used in the simulation data.}
    \label{measurement_noise_std}
    \scalebox{0.9}{
    \begin{tabular}{ccc}
    \hline
    \hline
    \textbf{Sensor}   &\textbf{Standard deviation}   &\textbf{Unit}    \\
    \hline
    $V_m$                        &$0.1$        &$[m/s]$         \\
    $\{\alpha, \beta\}_m$        &$0.1$        &$[deg]$         \\
    $\{G_x, G_y, G_z\}_m$        &$0.01$       &$[g]$       \\
    $\{p, q, r\}_m$              &$0.01$       &$[deg/s]$       \\
    $\{\psi, \theta, \phi\}_m$   &$0.01$       &$[deg]$         \\
    $\{x, y, z\}_m$              &$1$          &$[m]$         \\
    \hline
    \end{tabular}
    }
\end{table}

\begin{table}[h]
    \centering
    \caption{Aerospace sensors fault cases adopted in this paper.}
    \label{fault_cases}
        \scalebox{0.9}{
    \begin{tabular}{c|ccc}
    \hline
    \hline
    \textbf{Case}   &\textbf{Sensor}   &\textbf{Fault type}  &\textbf{Magnitude\textbf{*}}          \\
    \hline
    \textbf{9}   &$\{G_x, G_y, G_z\}_m$     &extra noise    &$0.1g\sim0.3g$             \\
    \textbf{8}   &$\{w_x, w_y, w_z\}_m$     &extra noise    &$5^o/s\sim10^o/s$              \\
    \textbf{7}   &$\{G_x, G_y, G_z\}_m$     &drift    &$\pm(0.1g\sim0.3g)$              \\
    \textbf{6}   &$\{w_x, w_y, w_z\}_m$     &drift    &$\pm(5^o/s\sim10^o/s)$              \\
    \textbf{5}   &$\beta_m$     &extra noise    &$5^o\sim10^o$              \\
    \textbf{4}   &$\beta_m$     &drift          &$\pm(5^o\sim10^o)$         \\
    \textbf{3}   &$\alpha_m$                &extra noise    &$5^o\sim10^o$              \\
    \textbf{2}   &$\alpha_m$                &drift          &$\pm(5^o\sim10^o)$         \\
    \textbf{1}   &$V_m$           &drift          &$-(50\%\sim100\%)$         \\   \hline
    \textbf{0}   &\multicolumn{3}{c}{clean measurement with noises and disturbances, no fault}  \\
    \hline
    \multicolumn{4}{l}{\textbf{*} Noise standard deviation and drift values defined in this column.}
    \end{tabular}
    }
\end{table}

\begin{figure}[h]
\centering
\includegraphics[trim={10pt 0pt 20pt 15pt},clip,width = .6\columnwidth]{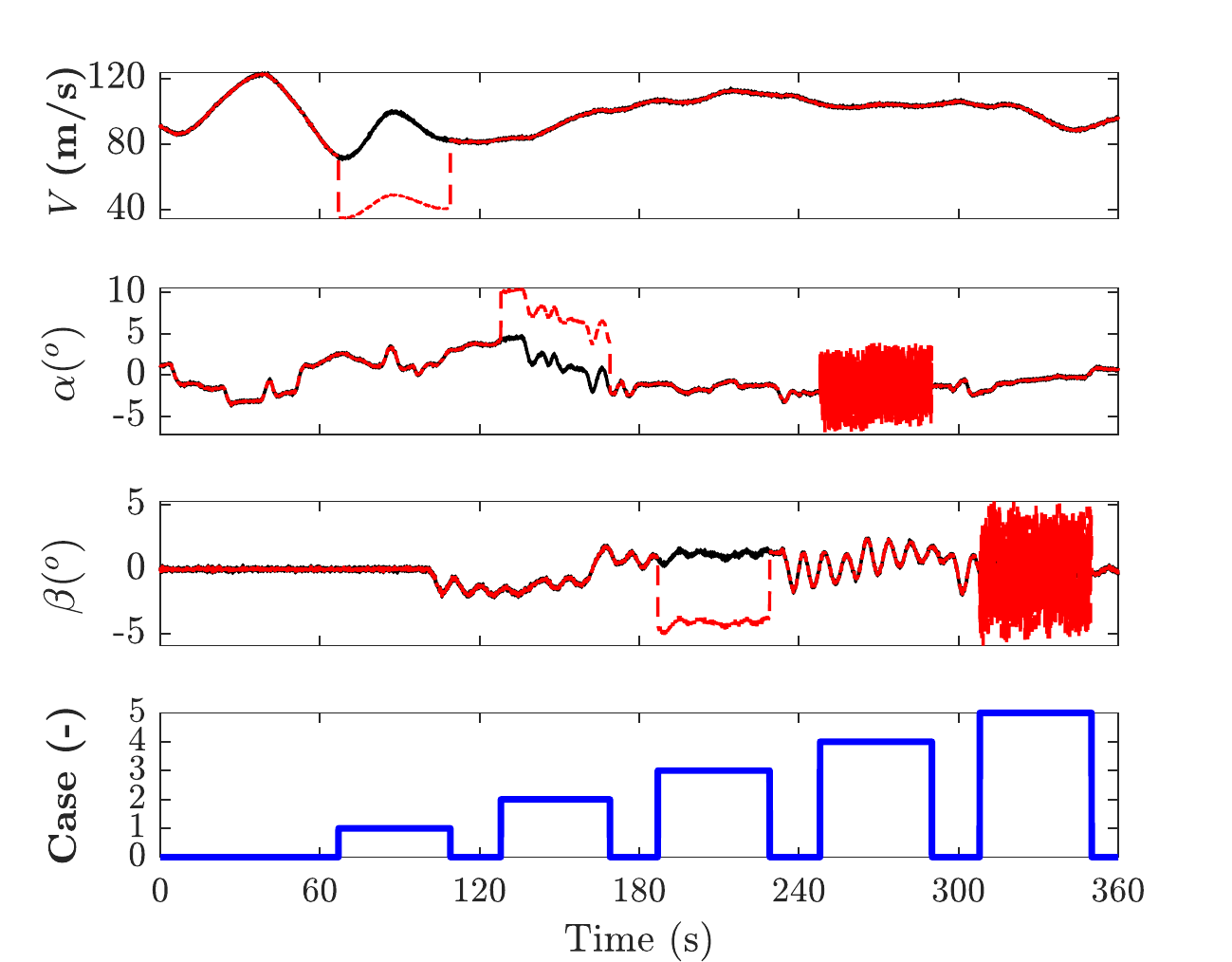}
\caption{Illustrative plot for the fault injection. Black lines denote clean states from real flight/simulation, red the fault-injected data. The FD scheme is expected to detect the different fault cases (blue thick line) based on available data measurements and proper DNN operations.}
\label{fault_injection_illustration}
\end{figure}

\subsection{ADS Fault Modeling and Injection}
Different sensor fault types have been discussed in previous works, which include ramp bias, oscillations, and drift. For airspeed, most flight accidents happened due to the Pitot tube being clogged by ice or rain. We thus consider drift fault for airspeed (measurement loss). For AOA and sideslip angles sensors, the deflection vanes may be stuck or perturbed by external atmosphere; we thus consider drift (constant bias) and extra noises faults. For IMU sensors, we model the faults following \cite{KF-27, KF-18, KF-AST}, and assume that drift/extra noise faults may appear for all rotational speeds/load factors simultaneously. As shown in Table \ref{fault_cases}, a total of 9 fault cases (for ADS and IMU sensors) are discussed, wherein the magnitude for each case is specified following \cite{EAAI}.  

We implement the aerospace sensors faults in an additive form; i.e., the ``clean" data (Case 0 in Table \ref{fault_cases}) are retrieved from real flight/simulations. Sensor faults are then injected into the measurement data. Following \cite{EAAI}, this injection is performed in a randomized manner; i.e., for every 60 seconds in the data, the fault cases occur randomly at randomized moments, with its duration (also randomized) not exceeding the 60 seconds. In Figure \ref{fault_injection_illustration}, different fault cases are injected to both airspeed, AOA, and sideslip angle for illustrative purposes.

\subsection{SDI for the Imagefication-based Intelligent FDC}

\begin{figure*}[t]
	\centering
	\includegraphics[width = \columnwidth]{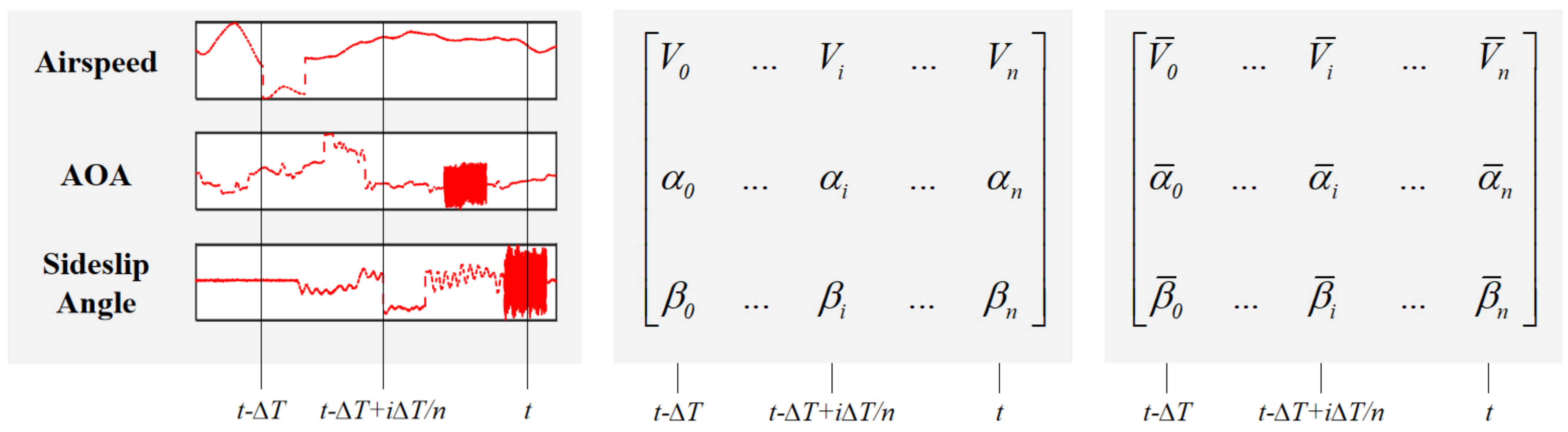}
	\centering
	\caption{Stacking of SDI for air data sensors. Left: flight records ($ V $, $ \alpha $, $ \beta $) are collected from real flights/simulations; faults are injected into the data, and a time window ($\Delta T$) is adopted to crop the data, which is then down-sampled to $1Hz$. Middle: the cropped and down-sampled data are concatenated as a matrix. Right: the matrix are linearly normalized  within the range of 0$\sim$1 along each row. The SDI stacking is performed for all 12 flight states and 3 load factors, yielding a $15\times31$ SDI grayscale image.}
	\label{SDI}
\end{figure*}

Imagefication-based intelligent FDC development hinges on the stacking of SDI, see Figure \ref{SDI}. Via simulations/real flights, measurement data from all the flight states/load factors are collected. We inject faults into the ADS/IMU measurement data, and stack them into a 2D matrix. In this matrix, each row stands for the historical measurements of a certain sensor (totally 15 sensors for 12 flight states and 3 load factors). The correlations of the states are then represented by the spatial coupling of different rows on the SDI. 

\begin{figure*}[t]
	\centering
	\includegraphics[width = \columnwidth]{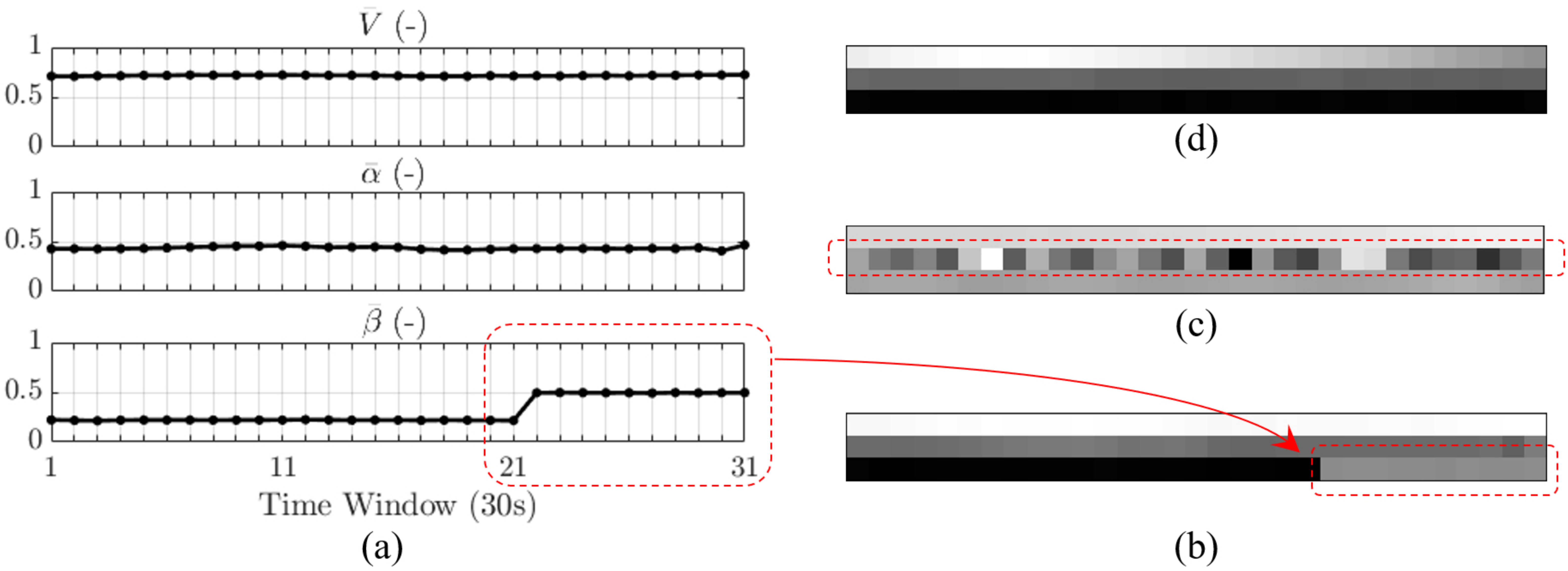}
	\centering
	\caption{An illustrative example of SDI; (a) the normalized sideslip angle ($\bar{\beta}$) drift fault occurs at 21s$\sim$31s; (b) the drift illustrates as a protruding strip on the stacked SDI; (c) illustration of AOA ($\bar{\alpha}$) sensor extra noise on the SDI, wherein the middle row oscillates with noises; (d) SDI for no-fault case, each row/column evolves as in Equations (\ref{translational_dynamics}$\sim$\ref{XYZ}), no abnormal region is found on the image.} 
	\label{SDI2objectdetection}
\end{figure*}

In stacking the SDI, time window is used to crop the data. At a certain moment $t$, we consider the states from $t-\Delta T$ to $t$ (both included), wherein $\Delta T=30s$\footnote{Following \cite{AST,EAAI}, this window may be understood as a compromise between the aircraft ``fast" motion modes of which the periods are in seconds (e.g. longitudinal short period, lateral roll), and ``slow" modes which typically last for tens/hundreds of seconds (e.g. longitudinal phugoid, Dutch roll).}. For different aircraft, the data are recorded in various sampling rates (e.g. 20$Hz$ for B$_1$, 30$Hz$ for B$_2$). We down-sample data to a unified frequency at $f_s=1/\Delta t$, wherein $\Delta t=\Delta T/30=1s$. We stack the SDI using the down-sampled sensors data. The SDI thus yields a size with 15 rows (12 flight states and 3 load factors) and 31 columns (31 down-sampled sensor measurement data). Apart from the spatial coupling between each rows on the SDI (representing the correlations of different states), sequential coupling of between different columns also captures the temporal features enclosed within the sensors measurement data.  

In the stacked SDI, the range of each state may vary significantly (e.g. angles in degrees, and position in kilometers). In practice, this may create numerical difficulties in the DNN training (singularities, error vanishing/explosion). Normalization thus is adopted. Following \cite{AST, EAAI}, this normalization is performed linearly (to a range between 0$\sim$1) along each row of the SDI. After normalization, the ``image" we acquire in the right plot of Figure \ref{SDI} is the SDI we adopted in our sensors FDC researches.

Pixel size of the SDI is $15\times31$, and elements enclosed within each index is normalized. The SDI thus similarizes the data format of a $15\times31$ sized grayscale image, see Figure \ref{SDI2objectdetection}. As in the figure, sensors fault illustrates as abnormal regions on the grayscale image (e.g. protruding stripes, oscillating noises). The original sensors FDC problem, therefore, is transformed into the abnormal object detection/classification on the grayscale SDI, wherein the input is a grayscale image (SDI), and output the fault labels (listed in Table \ref{fault_cases}). In the machine vision realm, the object detection/classification problem has long been a research topic wherein a plethora of mature methods/research outputs have appeared, which we can refer to in studying the sensors FDC problems.   

\section{VGG-based FDC for Aerospace Sensors}\label{Section_IV}
\subsection{Data Augmentation}
In previous section, the aerospace sensors FDC is transformed into an abnormal object detection problem on the SDI grayscale image. Pixel size of the SDI, however, is small. Based on the studies in \cite{dong2021deep, IJAE-robust}, this may create a bottleneck in training the FDC neural network. We thus propose to augment the SDI size, which on one hand provides a larger-sized basis for the convolutional kernels to extract features, and more specifically we hope the augmented size correspondent to VGG16 ($224\times224$) so that fine-tuning (from the VGG16 image classification neural network) may be adopted in training the sensors FDC neural network. 

Many data augmentation methods have appeared in literature which include interpolation, generative adversarial networks (GAN) \cite{shorten2019survey}, etc. Instinctively interpolation may be usable in our problem, as the SDI represents the down-sampled data of sensors measurement. However, interpolation along the (15) rows may render the drift fault (a protruding jump on the SDI) less obvious, and convolutional operations less effective. GAN-based data augmentation may also be usable in our case; but the GAN needs to be separately trained, which is still time-consuming.  

\begin{figure}[h]
	\centering
	\includegraphics[width = .8\textwidth]{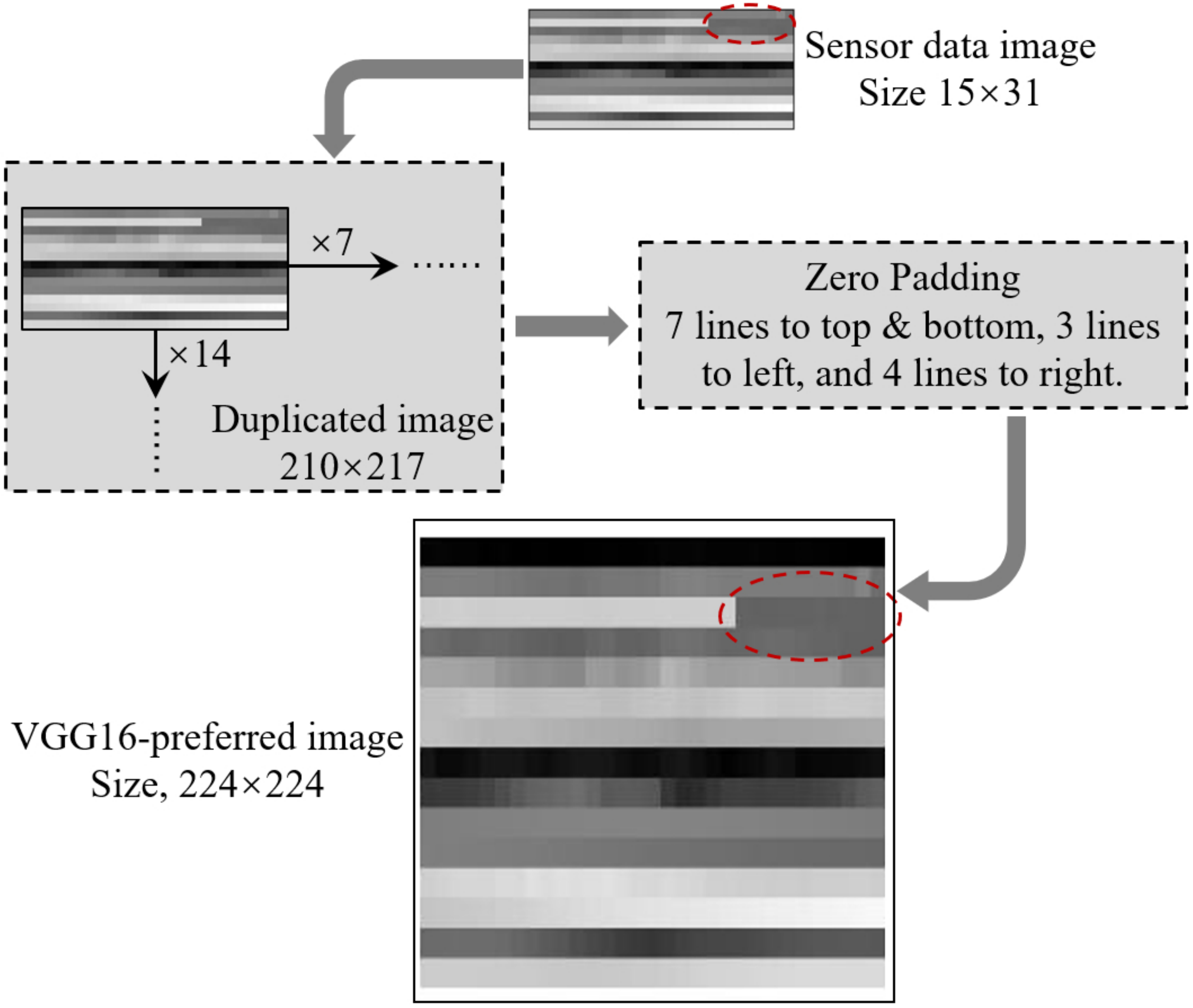}
	\centering
	\caption{Illustration of the proposed data augmentation method.}
	\label{fig:augment scheme}
\end{figure}

In our work we propose a ``duplicate" approach to augment the SDI size, as shown in  Figure \ref{fig:augment scheme}. Note that original size of the SDI is $15\times31$. It is then duplicated multiple times till the duplicated image falls closest (but still lower) to the VGG16-preferred size. Along the row dimension, we thus need to duplicate 7 times ($31\times7=217$), and along the column dimension 14 times is needed ($15\times14=210$). Around the duplicated image (size $210\times217$), we conduct zero-padding, i.e. to inflate its size with multiples rows/columns of 0 till it reaches $224\times224$ size. As shown in Figure \ref{fig:augment scheme}, we utilize 7 rows of 0 to both top and bottom of the duplicated image, 3 columns to left of the image, and 4 columns to right of the image. 

In our work we have investigated a total of 7 different methods to realize Figure \ref{fig:augment scheme}. Before presenting them we first define the 3 basic operations we utilized. We denote the SDI grayscale image as $M$, wherein the size of $M$ is $m (=15)$ by $n (=31)$. The first basic operation, \textit{Tile} is defined as in below equation:
\begin{equation}
{M_{dup}} = {\left( {\begin{array}{*{20}{c}}
M& \cdots &M\\
 \vdots & \ddots & \vdots \\
M& \cdots &M
\end{array}} \right)_{a \times b}}
	\label{eq:autmentation tile}
\end{equation}
wherein $a=14$ and $b=7$. In Equation (\ref{eq:autmentation tile}), the SDI matrix is repeated as a whole for a$\times$b times along the column and row directions, respectively. 

In the second basic operation \textit{Repeat}, every individual element in the SDI matrix is broadcasted separately, shown as below ($a=14$, $b=7$):
\begin{equation}
{M_{dup}} = \left( {\begin{array}{*{20}{c}}
{{{\left[ {\begin{array}{*{20}{c}}
{{M_{11}}}& \cdots &{{M_{11}}}\\
 \vdots & \ddots & \vdots \\
{{M_{11}}}& \cdots &{{M_{11}}}
\end{array}} \right]}_{a \times b}}}& \cdots & \cdots \\
 \vdots & \ddots & \vdots \\
 \vdots & \cdots &{{{\left[ {\begin{array}{*{20}{c}}
{{M_{mn}}}& \cdots &{{M_{mn}}}\\
 \vdots & \ddots & \vdots \\
{{M_{mn}}}& \cdots &{{M_{mn}}}
\end{array}} \right]}_{a \times b}}}
\end{array}} \right)
\label{eq:autmentation repeat}
\end{equation}

The third basic operation we defined for the data augmentation is \textit{Flip}. Consider the left-right flip matrix defined as:
\begin{equation}
	P_{lr}=\left(\begin{array}{ccccc}
		0 & 0 & \ldots & 0 & 1 \\
		0 & 0 & \ldots & 1 & 0 \\
		\vdots & \vdots & \iddots & \vdots & \vdots \\
		0 & 1 & \ldots & 0 & 0 \\
		1 & 0 & \ldots & 0 & 0
	\end{array}\right)_{n \times n}
	\label{eq:flip matrix}
\end{equation}
wherein the size of $P_{lr}$ is $n$-by-$n$. The left-right \textit{Flip} operation then yields:
\begin{equation}
	M_{lr}=MP_{lr}
	\label{eq:Mlr}
\end{equation}

Similarly, an up-down flip matrix is written as:
\begin{equation}
	P_{ud}=\left(\begin{array}{ccccc}
		0 & 0 & \ldots & 0 & 1 \\
		0 & 0 & \ldots & 1 & 0 \\
		\vdots & \vdots & \iddots & \vdots & \vdots \\
		0 & 1 & \ldots & 0 & 0 \\
		1 & 0 & \ldots & 0 & 0
	\end{array}\right)_{m \times m}
	\label{eq:flip matrix}
\end{equation}
wherein the size of $P_{ud}$ is $m$-by-$m$, the up-down \textit{Flip} operation yields:
\begin{equation}
	M_{ud}=P_{ud}M
	\label{eq:Mud}
\end{equation}

Given \textit{Tile}, \textit{Repeat}, and \textit{Flip} operations defined in Equations (\ref{eq:autmentation tile}$\sim$\ref{eq:Mud}), the 7 data augmentation methods we investigated are defined as following. 

\textbf{\textit{All\_Tile}}: This method is defined as in Equation (\ref{eq:autmentation tile}). The SDI ($M$) is repeated as a whole for a$\times$b times along the column and row directions, respectively.

\textbf{\textit{All\_Repeat}}: This method is shown as in Equation (\ref{eq:autmentation repeat}). Each individual element in the SDI is repeated for a$\times$b times along the column and row directions, separately.  

\textbf{\textit{All\_Flip}}: An intermediate matrix is firstly defined using the original SDI ($M$) and $M_{lr}$ as:
\begin{equation}
{M_i} = \left[ {M,{M_{lr}},M,{M_{lr}},M,{M_{lr}},M} \right]
\end{equation}
The up-down \textit{Flip} operation of $M_i$ yields $M_i{}_u{}_d$, the duplicated image is then written as:
\begin{equation}
    {M_{dup}} = \left[ {\underbrace {{M_i};{M_{iud}}; \cdots ;{M_i};{M_{iud}}}_{a}} \right]
\end{equation}

\textbf{\textit{LR\_Flip\_Tile}}: Left-right \textit{Flip} operation is firstly performed as in Equation (\ref{eq:Mlr}), \textit{Tile} operation is then adopted for $M_i$ (along column direction only) to realize $M_{dup}$.

\textbf{\textit{LR\_Flip\_Repeat}}:Left-right \textit{Flip} operation is firstly performed as in Equation (\ref{eq:Mlr}), \textit{Repeat} operation is then adopted for $M_i$ (along column direction only) to realize $M_{dup}$.

\textbf{\textit{UD\_Flip\_Tile}}: Up-down \textit{Flip} operation is firstly performed for $M$ to obtain $M_{ud}$, an intermediate matrix is defined as:
\begin{equation}
    {M_{i}} = \left[ {\underbrace {{M};{M_{ud}}; \cdots ;{M};{M_{ud}}}_{a}} \right]
    \label{eq:itermediate UD_Flip}
\end{equation}
\textit{Tile} operation is then performed for $M_{i}$ along the row direction to obtain $M_{dup}$.

\textbf{\textit{UD\_Flip\_Repeat}}: Up-down \textit{Flip} operation is firstly performed for $M$ to obtain $M_{ud}$, an intermediate matrix is defined as in Equation (\ref{eq:itermediate UD_Flip}), \textit{Tile} operation is then performed for $M_{i}$ along the row direction to obtain $M_{dup}$.

\subsection{VGG16 Fine-tuning for FDC}

VGG16 is fine-tuned directly in developing the aerospace sensors FDC neural network. An illustrative plot of the VGG16 architecture is shown in Figure \ref{figure_1}. In the previous chapter we have proposed 7 augmentation methods, we perform comparative studies on the 7 methods using the VGG16 fine-tuning results. Via each method, the SDI grayscale image (size $15\times31$) is inflated to VGG16-preferred size ($224\times224$). Taking the inflated image from each method, the pre-trained VGG16 (including both the architecture and internal parameters) is then loaded and fine-tuned using the same training data/algorithms/computational endeavors. The method that yields the best testing performance (same testing data is used for each method) is chosen as the method that we utilize, and the corresponding fine-tuned net adopted for further investigation (e.g. pruning in the next chapter).    

In the fine-tuning of VGG16, we establish the training environment via Python version 3.8 (in Windows 10), PyTorch version 1.9.0, and CUDA version 11.4. The training computer is also equipped with an Nvidia GeForce 3090 GPU, 64GB ram, and an i9-11900K CPU. Stochastic gradient descent (SGD) is adopted as the training algorithm (momentum 0.90), learning rate is set to a constant 1e-4, and batch size is set to 100 across all cases.

Based on the simulations/real flight data in Table \ref{aircraft_and_flightcondition}, a total of 111,965 SDI grayscale images are generated, which are then inflated with the data augmentation methods to VGG16-preferred size. All the images are also labeled as in Table \ref{fault_cases}. In the fine-tuning of VGG16 for each data augmentation method, 5-fold cross validation is adopted \cite{hong2021genetic}. We first evenly divide all the data into 5 folds (each fold 20\% with 22,393 images). We adopt each 1 fold of the images for testing, and the remaining 80\% for training. In such manner we are able to test the fine-tuned FDC net for 5 times separately (with each fold of the testing data strictly separated, but training data overlapped). We then combine the 5 testing results, and decide the best augmentation method with the best testing performance. 

\begin{figure}[htb]
	\centering
	\includegraphics[width = .8\textwidth]{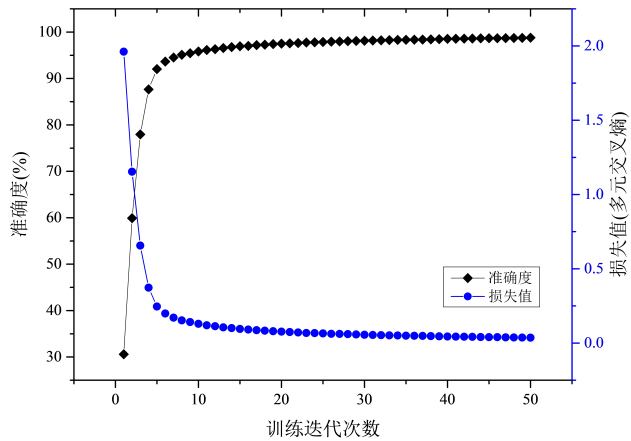}
	\centering
	\caption{Illustrative plots for the 5-fold cross validation training.}
	\label{fig:5-fold-illustration}
\end{figure}

Illustrative results of the 5-fold VGG16 fine-tuning are plotted in Figure \ref{fig:5-fold-illustration}. As shown in the figure, the FDC testing performance converges rapidly and steadily after only 50 epochs (each epoch costs 6.5min). We summarize the testing accuracy by averaging the testing results of the last 5 epochs for all the 5 fine-tuning processes in each method, and characterize the accuracy in Table \ref{7_augmentation_method_compare}; stand deviation (STD) of the 5-fold fine-tuning results are also listed in the table. As shown in the table, in the \textbf{\textit{All\_Repeat}} augmentation method, the 5-fold training claims the best overall testing performance (average highest, and STD lowest) at an average accuracy XX\%. Compared with previous works in \cite{AST} (80\%) and \cite{EAAI} (90\%), the results listed in Table \ref{7_augmentation_method_compare} are considered very promising.  

\begin{table*}[b]
    \centering
    \caption{Comparative studies on the 7 data augmentation methods}
    \label{7_augmentation_method_compare}
    \scalebox{0.68}{
    \begin{tabular}{l|ccccc|c}
    \hline
    \hline
     \textbf{\begin{tabular}[c]{@{}c@{}} Augmentation\\ Method\end{tabular}}   
    &\textbf{\begin{tabular}[c]{@{}c@{}} Accuracy\\ Fold 1 (\%)\end{tabular}}   
    &\textbf{\begin{tabular}[c]{@{}c@{}} Accuracy\\ Fold 2 (\%)\end{tabular}}    
    &\textbf{\begin{tabular}[c]{@{}c@{}} Accuracy\\ Fold 3 (\%)\end{tabular}}   
    &\textbf{\begin{tabular}[c]{@{}c@{}} Accuracy\\ Fold 4 (\%)\end{tabular}}   
    &\textbf{\begin{tabular}[c]{@{}c@{}} Accuracy\\ Fold 5 (\%)\end{tabular}} 
    &\textbf{\begin{tabular}[c]{@{}c@{}} Average, STD\\ (\%)\end{tabular}} 
    \\
    \hline
    %\xrowht[()]{3.5pt}
    \textbf{\textit{All\_Tile}}   
    &98.16  &98.51   &98.58   &98.57   &98.38     &98.44, 0.1756    \\  \hline  %\xrowht[()]{8pt}
    \textbf{\textit{All\_Repeat}}   
    &96.70  &97.04   &97.35   &97.39   &97.26     &97.15, 0.2847    \\  \hline     
    \textbf{\textit{All\_Flip}}   
    &97.61  &97.59   &98.02   &97.54   &97.76     &97.70, 0.1948    \\  \hline     
    \textbf{\textit{LR\_Flip\_Tile}}   
    &96.37  &96.94   &96.59   &96.85   &97.13     &96.78, 0.2988    \\  \hline     
    \textbf{\textit{LR\_Flip\_Repeat}}   
    &96.67  &97.72   &97.57   &97.72   &97.17     &97.37, 0.4514    \\  \hline     
    \textbf{\textit{UD\_Flip\_Tile}}   
    &97.44  &97.63   &97.78   &97.94   &97.89     &97.74, 0.2038    \\  \hline     
    \textbf{\textit{UD\_Flip\_Repeat}}   
    &97.50  &97.13   &97.20   &97.57   &97.85     &97.45, 0.2917    \\  \hline 
    \hline
    \end{tabular}
    }
\end{table*}

\subsection{Model Pruning for the Fine-tuned VGG16-FDC}

In the previous chapter, via fine-tuing and \textbf{\textit{All\_Repeat}} data augmentation method, the VGG16-based FDC net yields a best average testing accuracy at XX\%. We denote this net as VGG16-FDC, and visualize its internal operations in Figure X. In this figure, the features extracted in the 1st convolutional layer (totally 64 as following the VGG16 architecture in Figure \ref{figure_1}) are plotted. Note that many features are dark, which indicates the associated convolutional kernels not activated. Theoretically, although the \textbf{\textit{All\_Repeat}}-inflated image matches the size of VGG16, but the VGG16 architecture (e.g. 64 kernels in the 1st convolutional layer) was originally trained (optimized) for real images, which embraces much more complex pixel information as compared with the inflated SDI grayscale image (despite their sizes being the same). Although the VGG16 architecture is sufficient (as in Table \ref{7_augmentation_method_compare}) for the FDC problem, it may also be oversized. To truncate the less-relevant (e.g. inactivated kernels) operations from the fine-tuned VGG16 net, we perform model pruning. 

We have studied Random \cite{mittal2018recovering}, $L_1$ unstructured \cite{gao2020discover}, $L_2$ structured \cite{yuan2022membership}, and Taylor expansion based \cite{molchanov2016pruning} model pruning methods for the fine-tuned VGG16-FDC. For briefness issue, we detail the one with the best results, e.g. Tylor expansion pruning in this paper. Given the assumption that parameters enclosed within the DNN (e.g. weights in the VGG16-FDC) are independent, DNN model pruning is defined as an optimization problem with constraints shown in below:
\begin{equation}
	\begin{split}
		\min _{\mathcal{W}^{\prime}} \Delta\mathcal{C}(h_i)=:& \Big|\mathcal{C}\left(\mathcal{D} \mid \mathcal{W}^{\prime}\right)-\mathcal{C}(\mathcal{D} \mid \mathcal{W})\Big| \\
		=&\Big|\mathcal{C}(\mathcal{D},h_i=0)-\mathcal{C}(\mathcal{D},h_i)\Big|
		\quad \text { s.t. } \quad\left\|\mathcal{W}^{\prime}\right\|_{0} \leq B	
	\end{split}
\label{eq:pruning func}
\end{equation}
where $ h_i $ is the output produced from parameter $ i $, $ h = \{z^{(1)}_0,z^{(2)}_0,\cdots, z^{(c_{l})}_L\} $. $\mathcal{D}$ denotes training data $\{\mathcal{X}=\{x_0,\cdots,x_N\}$, $\mathcal{Y}=\{y_1,\cdots,y_N\}\}$, wherein $ x_* $ and $ y_* $ represent the net input and output, respectively.  $\mathcal{W}$ stands for the parameters to be optimized in DNN, i.e. $\left\{\left(\mathbf{w}_{1}^{1}, b_{1}^{1}\right), \cdots \left(\mathbf{w}_{L}^{C_{\ell}}, b_{L}^{C_{\ell}}\right)\right\}$. The optimization goal is to minimize a cost value $ \mathcal{C}(\mathcal{D} \mid \mathcal{W}) $. 
$ \mathcal{W}^{\prime} = \mathbf{g}\mathcal{W} $, wherein $ \mathbf{g} $ is the vectorized pruning gate $ \mathbf{g} \in\{0,1\}^{C_{l}} $. $ \mathcal{C}(\mathcal{D},h_i=0) $ the cost value if $ h_i $ is pruned, and $ \mathcal{C}(\mathcal{D},h_i) $ the cost value if $ h_i $ is not pruned.
The $ l_0 $ norm in $ \left\|\mathcal{W}^{\prime}\right\|_{0} \leq B $ bounds the number of non-zero parameters $ B $ in $ \mathcal{W'} $. Equation (\ref{eq:pruning func}) indicates that we refine a subset of parameters that preserves the accuracy of the network when pruning. The criteria adopted in the pruning is Taylor expansion. Approximating $ C(D, h_i = 0) $ with a first-order Taylor polynomial expansion at the neighbourhood of $ h_i = 0 $:
\begin{equation}
	\mathcal{C}\left(\mathcal{D}, h_{i}=0\right)=\mathcal{C}\left(\mathcal{D}, h_{i}\right)-\frac{\delta \mathcal{C}}{\delta h_{i}} h_{i}+R_{1}\left(h_{i}=0\right)
\end{equation}
wherein $ R_{1}(x) $ is the first-order expansion remainder. $ R_{1}(x) $ can be neglected due to the significant calculation required as well as a smaller second order term deduced from the ReLU activation function. Therefore $\Theta_{TE}:\mathbb{R}^{H_{l} \times W_{l} \times C_{l}} \rightarrow \mathbb{R}^{+}$ with
\begin{equation}
	\Theta_{T E}\left(h_{i}\right)=\left|\Delta \mathcal{C}\left(h_{i}\right)\right|=\left|\mathcal{C}\left(\mathcal{D}, h_{i}\right)-\frac{\delta \mathcal{C}}{\delta h_{i}} h_{i}-\mathcal{C}\left(\mathcal{D}, h_{i}\right)\right|=\left|\frac{\delta \mathcal{C}}{\delta h_{i}} h_{i}\right|
\end{equation}
and $ \Theta_{TE} $ is computed for a multi-variate output (e.g. a feature map extracted by a convolutional kernel) by
\begin{equation}
	\Theta_{T E}\left(z_{l}^{(k)}\right)=\left|\frac{1}{M} \sum_{m} \frac{\delta C}{\delta z_{l, m}^{(k)}} z_{l, m}^{(k)}\right|
\label{eq:prune func-end}
\end{equation}
wherein $ M $ is the length of the vectorized feature map.

We perform the model pruning algorithm defined in Equations (\ref{eq:pruning func}$\sim$\ref{eq:prune func-end}) to VGG16-FDC. The performance indexes we utilize in investigating the pruning process include number of parameters, model size, running time, and testing accuracy of the neural network. The pruning results (using Tylor expansion-based approach) are characterized in Table \ref{table:pruning}, wherein we denote the pruned net as pruned-VGG16-FDC. Note that in the table, the relative change is calculated by (After-Before)/Before. In the table, both number of parameters, size, and running time of the pruned net drops drastically whilst accuracy was slightly elevated. For accuracy improvement, this could be due to the pruning operation truncates redundant (less-relevant) structures from the original VGG16-FDC, simplifies the net architecture, hence renders the training data relatively more abundant (the net thus can be `better' trained). To summarize, based on Table \ref{table:pruning} the model pruning method we adopt yields an aerospace sensors FDC net at 5.60 MB size, 26 ms running time, and 98.90\% accuracy. 

\begin{table}[htb]
	%\centering
	\caption{Model pruning results of the fine-tuned VGG16-FDC neural network.}
	\label{table:pruning}
	\begin{tabular}{ccccc}
		\hline
		\hline
		&\textbf{Parameters}  & \textbf{ Size (MB)}   & \textbf{Time (ms)} 
		& \textbf{Accuracy (\%)} \\  
		\hline
	Before	& 134300362    		&   512     & 315  & 97.37 \\
	After	& 1387333       	&   5.60    & 26   & 98.90 \\
	\hline
\makecell[c]{Relative \\ Change}	&  98.97\% $\downarrow$	&  98.91\% $\downarrow$ &  91.75\% $\downarrow$ & 1.53\% $\uparrow$	\\
		\hline
	\end{tabular}
\end{table}

\section{Explainability Analysis of the FDC Net}\label{Section_V}

%\subsection{Explainability Analysis}
To better visualize the internal operations enclosed within the FDC neural network (i.e. the pruned-VGG16-FDC), we adopt Gradient-weighted Class Activation Mapping (Grad-CAM \cite{selvaraju2017}]), which is a variant of the original CAM method. Suppose Grad-CAM $L_{\text {Grad-CAM }}^{c} \in \mathbb{R}^{u \times v}$ with width $ u $ and height $ v $ for any class $ c $, $\frac{\partial y^{c}}{\partial A^{k}}$ is the gradient of the score for class $ c $, $ y^{c} $ with respect to feature map activations $ A^{k} $ of a convolutional layer. These gradients flowing back are global-average-pooled 2 over the width and height dimensions (indexed by $ i $ and $ j $ respectively) to obtain the neuron importance weights $ \alpha^{c}_{k} $:
\begin{equation}
\alpha_{k}^{c}=\hspace{-0.5cm}\overbrace{\frac{1}{Z} \sum_{i} \sum_{j}}^{\text{global average pooling}} \hspace{-1.5cm} \underbrace{\frac{\partial y^{c}}{\partial A_{i j}^{k}}}_{\text{gradients via backprop}}	
\end{equation}
wherein Z is the number of pixels in the feature map ($ Z = \sum_{i}\sum_{j}1 $). The weight $ \alpha^{c}_{k} $  represents a partial linearization of the deep network downstream from A, and captures the ``attention" of feature map $ k $ for a target class $ c $. Finally, the Grad-CAM can be obtained by the sequential processing  of a linear combination (weighted combination of forward activation maps) and a nonlinear operation (ReLU function):
\begin{equation}
	L_{\text {Grad-CAM }}^{c}=\operatorname{ReLU} \underbrace{\left(\sum_{k} \alpha_{k}^{c} A^{k}\right)}_{\text {linear combination }}
\end{equation}

\begin{figure}[h]
	\centering
	\includegraphics[width = .8\textwidth]{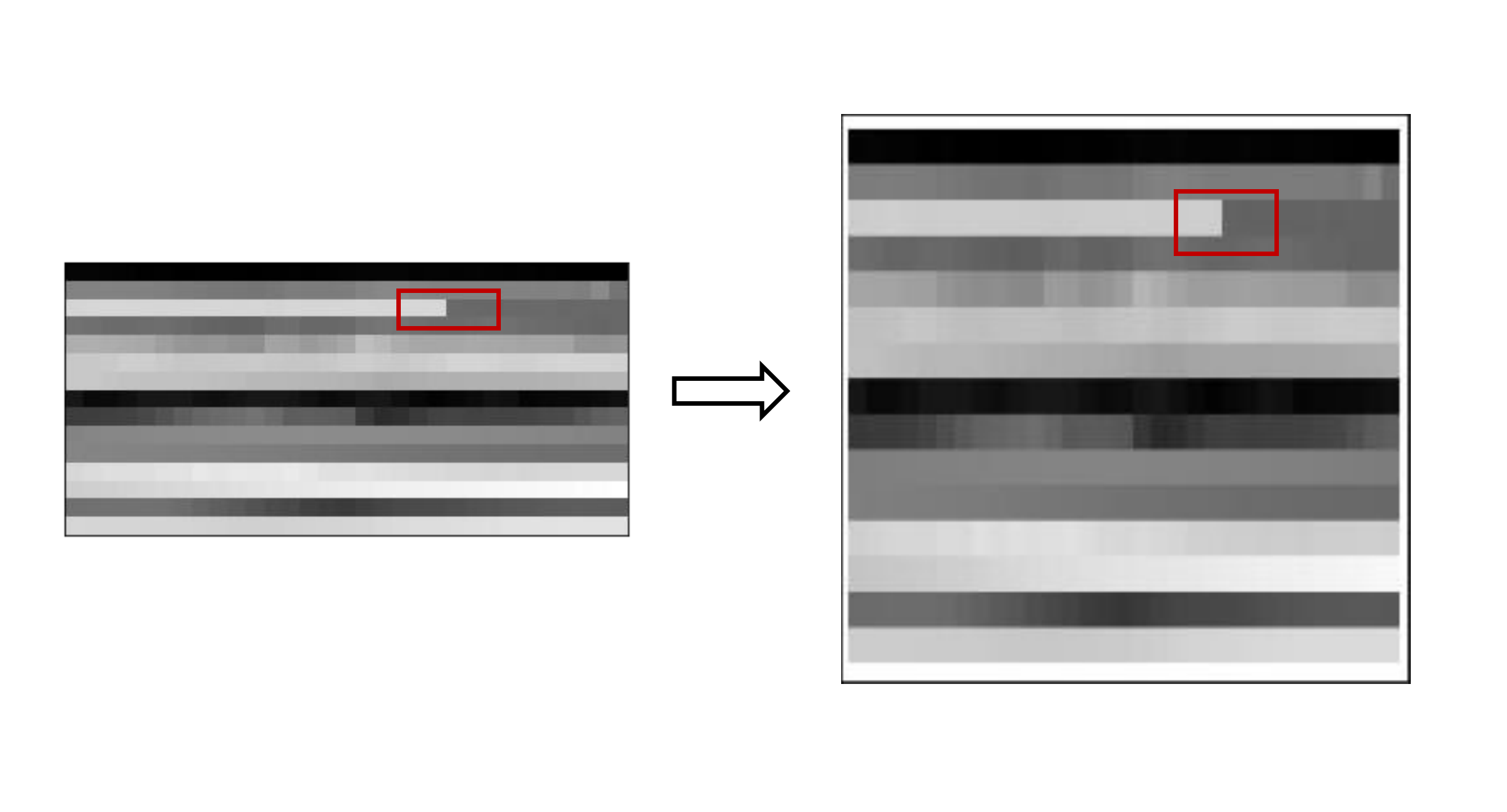}
	\caption{SDI and inflated images for a $\beta$ sensor fault, fault is marked in red (images not in scale).}
	\label{fig:before_after_aug}
\end{figure}

\begin{figure}[!h]
	\centering
	\includegraphics[width =1.0\textwidth,trim=100 0 100 0]{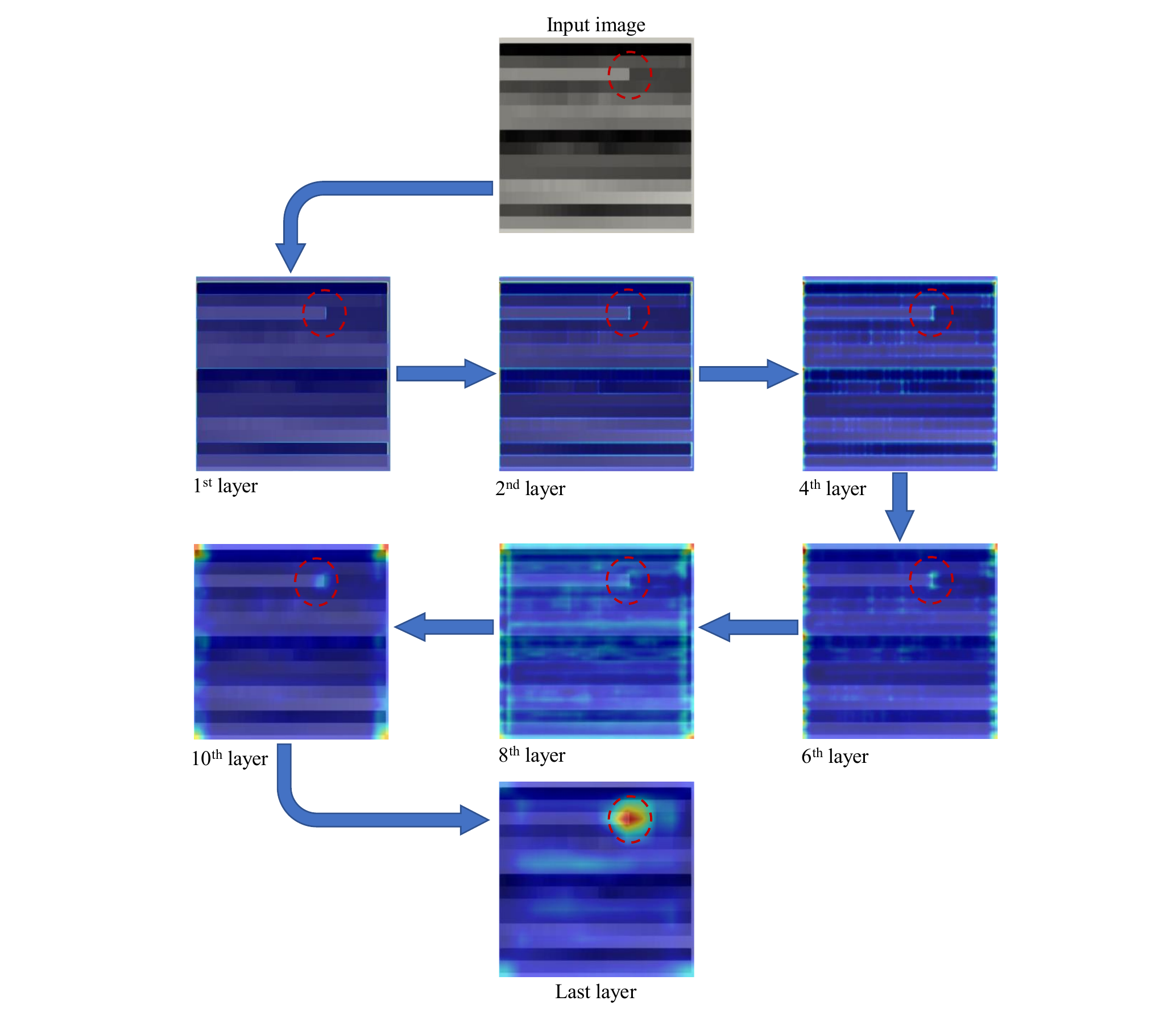}
	\caption{Illustrative Grad-CAM results of the pruned-VGG16-FDC neural network; fault is marked in red.}
	\label{fig:gradcam}
\end{figure}

Figures \ref{fig:before_after_aug}$\sim$\ref{fig:gradcam} are illustrative results of the Grad-CAM analysis. In Figure \ref{fig:before_after_aug}, a sideslip angle drift fault is injected to the sensors as in Figure \ref{SDI2objectdetection}. On the original SDI image (size $15\times31$), the drift fault is marked in red as a protruding strip. After the \textbf{\textit{All\_Repeat}} augmentation method, the image is inflated to VGG16-preferred size (fault also marked in red). Using the inflated image shown in Figure \ref{fig:before_after_aug} as input, Grad-CAM results from different layers of the pruned-VGG16-FDC neural network are plotted in Figure \ref{fig:gradcam}. In the first several layers of the net (1st and 2nd), the convolution focuses on extracting low-level features such as edges/lines on the image. As the convolutional layers stack deeper, more sophisticated features appear in the 6th and 10th layers, and the ``hotter" area illustrated on the Grad-CAM plots reveals the ``attention" (on the input image) where the associated convolutional layer is focused on. As shown in Figure \ref{fig:gradcam}, the hotter areas in the 6th and 10th layers correspond to the protruding strip shown on the net input. In the last layer of the neural network (13th), the hotter area focuses exclusively around the fault region. As shown in Figure \ref{fig:gradcam}, the CAM plots correspond to the abnormal region wherein the fault occurs, the operations enclosed within the pruned-VGG16-FDC neural network is therefore considered explainable. 

%\subsection{Real Flights-based Test}

\section{Conclusions}\label{Section_VI}

Following the recent developments of imagefication-based intelligent fault detection and diagnosis researches, a VGG16-based fault detection and classification (FDC) framework is proposed for aerospace sensors. Measurements of the sensors are first stacked into a grayscale image; data augmentation is then studied to inflate the stacked image to the size that is compatible with VGG16, a pre-trained object classification deep neural network in the machine vision realm. Detection and classification of the sensors faults are transformed into abnormal regions detection problem on the inflated image. The sensors FDC neural network is trained via fine-tuning the VGG16, which is further pruned for model compression. Via extensive tests, the proposed net in this paper yields a 98.90\% FDC accuracy across 4 aircraft at 5 different flight conditions. We also perform explainability analysis of the proposed net via Grad-CAM (commonly used in the machine vision realm), which yields promising results.

\section*{Acknowledgement}
We are indebted to Dr. Hang Zhao for the discussions on this paper. 

\bibliographystyle{elsarticle-num}
\bibliography{ref}

\begin{thebibliography}{10}
\expandafter\ifx\csname url\endcsname\relax
  \def\url#1{\texttt{#1}}\fi
\expandafter\ifx\csname urlprefix\endcsname\relax\def\urlprefix{URL }\fi
\expandafter\ifx\csname href\endcsname\relax
  \def\href#1#2{#2} \def\path#1{#1}\fi

\bibitem{X31_crash}
J.~Levine,
  \href{http://www.nasa.gov/centers/dryden/news/X-Press/stories/2004/013004/new_x31.html}{X-31's
  loss} (2004 (accessed September 21, 2020)).
\newline\urlprefix\url{http://www.nasa.gov/centers/dryden/news/X-Press/stories/2004/013004/new_x31.html}

\bibitem{AF330}
F.~Bureau d’Enquetes et d’Analyses pour la securite de~l’aviation civile,
  Paris, \href{https://www.bea.aero/docspa/2009/
  f-cp090601e3.en/pdf/f-cp090601e3.en.pdf}{Report 3: on the accident on 1 june
  2009 to the airbus a330–203 registered f-gzcp operated by air france flight
  af 447 rio de janeiro–paris} (2011 (accessed September 21, 2020)).
\newline\urlprefix\url{https://www.bea.aero/docspa/2009/
  f-cp090601e3.en/pdf/f-cp090601e3.en.pdf}

\bibitem{B737Max}
FAA, \href{https://www.faa.gov/news/updates/?newsId=93206}{FAA Updates on
  Boeing 737 MAX} (2020 (accessed 7 October, 2020)).
\newline\urlprefix\url{https://www.faa.gov/news/updates/?newsId=93206}

\bibitem{HR-1}
P.~Goupil, J.~Boada-Bauxell, A.~Marcos, E.~Cortet, M.~Kerr, H.~Costa, Airbus
  efforts towards advanced real-time fault diagnosis and fault tolerant
  control, IFAC Proceedings Volumes 47~(3) (2014) 3471--3476.

\bibitem{HR-2}
E.~Dubrova, Fault-tolerant design, Springer, 2013.

\bibitem{HR-3}
P.~Goupil, Airbus state of the art and practices on {FDI} and {FTC} in flight
  control system, Control Engineering Practice 19~(6) (2011) 524--539.

\bibitem{MB-review}
J.~Marzat, H.~Piet-Lahanier, F.~Damongeot, E.~Walter, Model-based fault
  diagnosis for aerospace systems: a survey, Proceedings of the Institution of
  Mechanical Engineers, Part G: Journal of Aerospace Engineering 226~(10)
  (2012) 1329--1360.

\bibitem{own_IEEE_TAES}
Y.~Dong, J.~Tao, Y.~Zhang, W.~Lin, J.~Ai, Deep learning in aircraft design,
  dynamics, and control: Review and prospects, IEEE Transactions on Aerospace
  and Electronic Systems (2021).

\bibitem{ablationstudy}
R.~Meyes, M.~Lu, C.~W. de~Puiseau, T.~Meisen, Ablation studies in artificial
  neural networks, arXiv preprint arXiv:1901.08644 (2019).

\bibitem{zhou2016}
B.~Zhou, A.~Khosla, A.~Lapedriza, A.~Oliva, A.~Torralba, Learning deep features
  for discriminative localization, in: Proceedings of the IEEE conference on
  computer vision and pattern recognition, 2016, pp. 2921--2929.

\bibitem{transferlearning}
M.~Long, J.~Wang, Learning transferable features with deep adaptation networks,
  JMLR.org (2015).

\bibitem{AST}
Y.~Dong, An application of deep neural networks to the in-flight parameter
  identification for detection and characterization of aircraft icing,
  Aerospace Science and Technology 77 (2018) 34--49.

\bibitem{EAAI}
Y.~Dong, Implementing deep learning for comprehensive aircraft icing and
  actuator/sensor fault detection/identification, Engineering Applications of
  Artificial Intelligence 83 (2019) 28--44.

\bibitem{IJAE-yiming}
Y.~Zhang, H.~Zhao, J.~Ma, Y.~Zhao, Y.~Dong, J.~Ai, A deep neural network-based
  fault detection scheme for aircraft imu sensors, International Journal of
  Aerospace Engineering 2021 (2021).

\bibitem{dong2021deep}
Y.~Dong, J.~Wen, Y.~Zhang, J.~Ai, Deep neural networks-based air data sensors
  fault detection for aircraft, in: 2021 33rd Chinese Control and Decision
  Conference (CCDC), IEEE, 2021, pp. 442--447.

\bibitem{IJAE-robust}
Y.~Zhao, H.~Zhao, J.~Ai, Y.~Dong, Robust data-driven fault detection: An
  application to aircraft air data sensors, International Journal of Aerospace
  Engineering 2022 (2022).

\bibitem{KF-27}
L.~Van~Eykeren, Q.~Chu, Sensor fault detection and isolation for aircraft
  control systems by kinematic relations, Control Engineering Practice 31
  (2014) 200--210.

\bibitem{KF-18}
M.~Ariola, M.~Mattei, I.~Notaro, F.~Corraro, A.~Sollazzo, An {SFDI}
  observer--based scheme for a general aviation aircraft, International Journal
  of Applied Mathematics and Computer Science 25~(1) (2015) 149--158.

\bibitem{KF-AST}
Q.~He, W.~Zhang, P.~Lu, J.~Liu, Performance comparison of representative
  model-based fault reconstruction algorithms for aircraft sensor fault
  detection and diagnosis, Aerospace Science and Technology 98 (2020) 105649.

\bibitem{KF-11}
P.~Lu, L.~Van~Eykeren, E.-J. Van~Kampen, Q.~P. Chu, B.~Yu, Adaptive hybrid
  unscented {K}alman filter for aircraft sensor fault detection, isolation and
  reconstruction, in: AIAA Guidance, Navigation, and Control Conference, 2014,
  p. 1145.

\bibitem{KF-7}
P.~Lu, L.~Van~Eykeren, E.~Van~Kampen, C.~De~Visser, Q.~Chu, Adaptive three-step
  {K}alman filter for air data sensor fault detection and diagnosis, Journal of
  Guidance, Control, and Dynamics 39~(3) (2016) 590--604.

\bibitem{Hinf}
P.~Freeman, P.~Seiler, G.~J. Balas, Air data system fault modeling and
  detection, Control Engineering Practice 21~(10) (2013) 1290--1301.

\bibitem{Hinf-1}
M.~Mattei, G.~Paviglianiti, Managing sensor hardware redundancy on a small
  commercial aircraft with {H}$\infty$ {FDI} observers, IFAC Proceedings
  Volumes 38~(1) (2005) 347--352.

\bibitem{Hinf-2}
F.~Amato, C.~Cosentino, M.~Mattei, G.~Paviglianiti, A direct/functional
  redundancy scheme for fault detection and isolation on an aircraft, Aerospace
  Science and Technology 10~(4) (2006) 338--345.

\bibitem{AST-robust}
Y.~Xue, Z.~Zhen, L.~Yang, L.~Wen, Adaptive fault-tolerant control for
  carrier-based uav with actuator failures, Aerospace Science and Technology
  107 (2020) 106227.

\bibitem{MHE-1}
Y.~Wan, T.~Keviczky, Real-time fault-tolerant moving horizon air data
  estimation for the reconfigure benchmark, IEEE Transactions on Control
  Systems Technology 27~(3) (2018) 997--1011.

\bibitem{MHE-2}
Y.~Wan, T.~Keviczky, M.~Verhaegen, Robust air data sensor fault diagnosis with
  enhanced fault sensitivity using moving horizon estimation, in: 2016 American
  Control Conference (ACC), IEEE, 2016, pp. 5969--5975.

\bibitem{MHE-3}
Y.~Wan, T.~Keviczky, Implementation of real-time moving horizon estimation for
  robust air data sensor fault diagnosis in the reconfigure benchmark,
  IFAC-PapersOnLine 49~(17) (2016) 64--69.

\bibitem{NGA-1}
P.~Castaldi, N.~Mimmo, S.~Simani, Avionic air data sensors fault detection and
  isolation by means of singular perturbation and geometric approach, Sensors
  17~(10) (2017) 2202.

\bibitem{NGA-2}
P.~Castaldi, W.~Geri, M.~Bonfe, S.~Simani, M.~Benini, Design of residual
  generators and adaptive filters for the {FDI} of aircraft model sensors,
  Control Engineering Practice 18~(5) (2010) 449--459.

\bibitem{AST-barrier}
X.~Zhu, J.~Chen, Z.~H. Zhu, Adaptive learning observer for spacecraft attitude
  control with actuator fault, Aerospace Science and Technology (2020) 106389.

\bibitem{SVO}
P.~Rosa, C.~Silvestre, Fault detection and isolation of {LPV} systems using
  set-valued observers: An application to a fixed-wing aircraft, Control
  Engineering Practice 21~(3) (2013) 242--252.

\bibitem{NN-26}
S.~Hussain, M.~Mokhtar, J.~M. Howe, Sensor failure detection, identification,
  and accommodation using fully connected cascade neural network, IEEE
  Transactions on Industrial Electronics 62~(3) (2014) 1683--1692.

\bibitem{NN-29}
S.~Hussain, M.~Mokhtar, J.~M. Howe, Aircraft sensor estimation for fault
  tolerant flight control system using fully connected cascade neural network,
  in: The 2013 International Joint Conference on Neural Networks (IJCNN), IEEE,
  2013, pp. 1--8.

\bibitem{NN-17}
L.~Garbarino, G.~Zazzaro, N.~Genito, G.~Fasano, D.~Accardo, Neural network
  based architecture for fault detection and isolation in air data systems, in:
  2013 IEEE/AIAA 32nd Digital Avionics Systems Conference (DASC), IEEE, 2013,
  pp. 2D4--1.

\bibitem{NN-19}
S.~Gururajan, M.~L. Fravolini, H.~Chao, M.~Rhudy, M.~R. Napolitano, Performance
  evaluation of neural network based approaches for airspeed sensor failure
  accommodation on a small {UAV}, in: 21st Mediterranean Conference on Control
  and Automation, IEEE, 2013, pp. 603--608.

\bibitem{MJ-4}
M.~Kordestani, M.~F. Samadi, M.~Saif, A distributed fault detection and
  isolation method for multifunctional spoiler system, in: 2018 IEEE 61st
  International Midwest Symposium on Circuits and Systems (MWSCAS), IEEE, 2018,
  pp. 380--383.

\bibitem{NN-24}
A.~Abbaspour, P.~Aboutalebi, K.~K. Yen, A.~Sargolzaei, Neural adaptive
  observer-based sensor and actuator fault detection in nonlinear systems:
  Application in {UAV}, ISA {T}ransactions 67 (2017) 317--329.

\bibitem{NN-31}
M.~L. Fravolini, M.~R. Napolitano, G.~Del~Core, U.~Papa, Experimental interval
  models for the robust fault detection of aircraft air data sensors, Control
  Engineering Practice 78 (2018) 196--212.

\bibitem{DataDriven-1}
M.~L. Fravolini, G.~Del~Core, U.~Papa, P.~Valigi, M.~R. Napolitano, Data-driven
  schemes for robust fault detection of air data system sensors, IEEE
  Transactions on Control Systems Technology 27~(1) (2017) 234--248.

\bibitem{DataDriven-3}
M.~L. Fravolini, M.~Rhudy, S.~Gururajan, S.~Cascianelli, M.~Napolitano,
  Experimental evaluation of two pitot free analytical redundancy techniques
  for the estimation of the airspeed of an {UAV}, SAE International Journal of
  Aerospace 7~(2014-01-2163) (2014) 109--116.

\bibitem{DataDriven-4}
S.~Gururajan, M.~L. Fravolini, H.~Chao, M.~Rhudy, M.~R. Napolitano, Performance
  evaluation of neural network based approaches for airspeed sensor failure
  accommodation on a small {UAV}, in: 21st Mediterranean Conference on Control
  and Automation, IEEE, 2013, pp. 603--608.

\bibitem{deeplearningbook}
{I. Goodfellow, A. Courville, and Y. Bengio}, Deep learning, Vol.~1, MIT press
  Cambridge, 2016.

\bibitem{RNN-1}
H.~A. Talebi, K.~Khorasani, S.~Tafazoli, A recurrent neural-network-based
  sensor and actuator fault detection and isolation for nonlinear systems with
  application to the satellite's attitude control subsystem, IEEE
  {T}ransactions on Neural Networks 20~(1) (2008) 45--60.

\bibitem{RNN-2}
M.~Chen, P.~Shi, C.-C. Lim, Adaptive neural fault-tolerant control of a 3-{DOF}
  model helicopter system, IEEE Transactions on Systems, Man, and Cybernetics:
  Systems 46~(2) (2015) 260--270.

\bibitem{RNN-3}
E.~Sobhani-Tehrani, H.~A. Talebi, K.~Khorasani, Hybrid fault diagnosis of
  nonlinear systems using neural parameter estimators, Neural Networks 50
  (2014) 12--32.

\bibitem{MJ-3}
L.~Chen, J.~Cao, K.~Wu, Z.~Zhang, Application of generalized frequency response
  functions and improved convolutional neural network to fault diagnosis of
  heavy-duty industrial robot, Robotics and Computer-Integrated Manufacturing
  73 (2022) 102228.

\bibitem{MJ-7}
R.~M. Souza, E.~G. Nascimento, U.~A. Miranda, W.~J. Silva, H.~A. Lepikson, Deep
  learning for diagnosis and classification of faults in industrial rotating
  machinery, Computers \& Industrial Engineering 153 (2021) 107060.

\bibitem{simonyan2014very}
K.~Simonyan, A.~Zisserman, Very deep convolutional networks for large-scale
  image recognition, arXiv preprint arXiv:1409.1556 (2014).

\bibitem{deng2009imagenet}
J.~Deng, W.~Dong, R.~Socher, L.-J. Li, K.~Li, L.~Fei-Fei, Imagenet: A
  large-scale hierarchical image database, in: 2009 IEEE conference on computer
  vision and pattern recognition, Ieee, 2009, pp. 248--255.

\bibitem{YOLO-1}
J.~Redmon, S.~Divvala, R.~Girshick, A.~Farhadi, You only look once: Unified,
  real-time object detection, in: Proceedings of the IEEE conference on
  computer vision and pattern recognition, 2016, pp. 779--788.

\bibitem{YOLO-2}
J.~Redmon, A.~Farhadi, Yolo9000: better, faster, stronger, in: Proceedings of
  the IEEE conference on computer vision and pattern recognition, 2017, pp.
  7263--7271.

\bibitem{YOLO-3}
J.~Redmon, A.~Farhadi, Yolov3: An incremental improvement, arXiv preprint
  arXiv:1804.02767 (2018).

\bibitem{fast-rcnn}
R.~Girshick, Fast r-cnn, in: Proceedings of the IEEE international conference
  on computer vision, 2015, pp. 1440--1448.

\bibitem{faster-rcnn}
S.~Ren, K.~He, R.~Girshick, J.~Sun, Faster r-cnn: Towards real-time object
  detection with region proposal networks, Advances in neural information
  processing systems 28 (2015).

\bibitem{mask-rcnn}
K.~He, G.~Gkioxari, P.~Doll{\'a}r, R.~Girshick, Mask r-cnn, in: Proceedings of
  the IEEE international conference on computer vision, 2017, pp. 2961--2969.

\bibitem{denil2013predicting}
M.~Denil, B.~Shakibi, L.~Dinh, M.~Ranzato, N.~De~Freitas, Predicting parameters
  in deep learning, Advances in neural information processing systems 26
  (2013).

\bibitem{luo2017thinet}
J.-H. Luo, J.~Wu, W.~Lin, Thinet: A filter level pruning method for deep neural
  network compression, in: Proceedings of the IEEE international conference on
  computer vision, 2017, pp. 5058--5066.

\bibitem{liu2017learning}
Z.~Liu, J.~Li, Z.~Shen, G.~Huang, S.~Yan, C.~Zhang, Learning efficient
  convolutional networks through network slimming, in: Proceedings of the IEEE
  international conference on computer vision, 2017, pp. 2736--2744.

\bibitem{li2016pruning}
H.~Li, A.~Kadav, I.~Durdanovic, H.~Samet, H.~P. Graf, Pruning filters for
  efficient convnets, arXiv preprint arXiv:1608.08710 (2016).

\bibitem{JAIS}
Y.~Dong, Deep learning-based opponent aircraft attitude detection in autonomous
  air combat, Journal of Aerospace Information Systems 16~(4) (2019) 162--167.

\bibitem{selvaraju2017}
R.~R. Selvaraju, M.~Cogswell, A.~Das, R.~Vedantam, D.~Parikh, D.~Batra,
  Grad-{CAM}: Visual explanations from deep networks via gradient-based
  localization, in: Proceedings of the IEEE international conference on
  computer vision, 2017, pp. 618--626.

\bibitem{chattopadhay2018}
A.~Chattopadhay, A.~Sarkar, P.~Howlader, V.~N. Balasubramanian, Grad-{CAM}++:
  Generalized gradient-based visual explanations for deep convolutional
  networks, in: 2018 IEEE winter conference on applications of computer vision
  (WACV), IEEE, 2018, pp. 839--847.

\bibitem{omeiza2019}
D.~Omeiza, S.~Speakman, C.~Cintas, K.~Weldermariam, Smooth {G}rad-{CAM}++: An
  enhanced inference level visualization technique for deep convolutional
  neural network models, arXiv preprint arXiv:1908.01224 (2019).

\bibitem{wang2020}
H.~Wang, Z.~Wang, M.~Du, F.~Yang, Z.~Zhang, S.~Ding, P.~Mardziel, X.~Hu,
  Score-{CAM}: score-weighted visual explanations for convolutional neural
  networks, in: Proceedings of the IEEE/CVF conference on computer vision and
  pattern recognition workshops, 2020, pp. 24--25.

\bibitem{wang2020ss}
H.~Wang, R.~Naidu, J.~Michael, S.~S. Kundu, S{S}-{CAM}: smoothed score-{CAM}
  for sharper visual feature localization, arXiv preprint arXiv:2006.14255
  (2020).

\bibitem{ramaswamy2020}
H.~G. Ramaswamy, et~al., Ablation-{CAM}: Visual explanations for deep
  convolutional network via gradient-free localization, in: Proceedings of the
  IEEE/CVF Winter Conference on Applications of Computer Vision, 2020, pp.
  983--991.

\bibitem{dong2016full}
Y.~Dong, Y.~Zhang, J.~Ai, Full-altitude attitude angles envelope and model
  predictive control-based attitude angles protection for civil aircraft,
  Aerospace Science and Technology 55 (2016) 292--306.

\bibitem{hohndorf2017reconstruction}
L.~H{\"o}hndorf, J.~Siegel, J.~Sembiring, P.~Koppitz, F.~Holzapfel,
  Reconstruction of aircraft states during landing based on quick access
  recorder data, Journal of Guidance, Control, and Dynamics 40~(9) (2017)
  2393--2398.

\bibitem{nelson1998flight}
R.~C. Nelson, et~al., Flight stability and automatic control, Vol.~2,
  WCB/McGraw Hill New York, 1998.

\bibitem{MIL1797}
{Department of Defense},
  \href{https://cafe.foundation/v2/pdf_tech/Flying.Qualities/PAV.FlyQual.Mil1797A.pdf}{Flying
  Qualities of Piloted Aircraft} (1997 (accessed October 20, 2020)).
\newline\urlprefix\url{https://cafe.foundation/v2/pdf_tech/Flying.Qualities/PAV.FlyQual.Mil1797A.pdf}

\bibitem{shorten2019survey}
C.~Shorten, T.~M. Khoshgoftaar, A survey on image data augmentation for deep
  learning, Journal of Big Data 6~(1) (2019) 1--48.

\bibitem{hong2021genetic}
D.~Hong, Y.-Y. Zheng, Y.~Xin, L.~Sun, H.~Yang, M.-Y. Lin, C.~Liu, B.-N. Li,
  Z.-W. Zhang, J.~Zhuang, et~al., Genetic syndromes screening by facial
  recognition technology: Vgg-16 screening model construction and evaluation,
  Orphanet Journal of Rare Diseases 16~(1) (2021) 1--8.

\bibitem{mittal2018recovering}
D.~Mittal, S.~Bhardwaj, M.~M. Khapra, B.~Ravindran, Recovering from random
  pruning: On the plasticity of deep convolutional neural networks, in: 2018
  IEEE Winter Conference on Applications of Computer Vision (WACV), IEEE, 2018,
  pp. 848--857.

\bibitem{gao2020discover}
S.~Gao, A discover of class and image level variance between different pruning
  methods on convolutional neural networks, in: 2020 IEEE International
  Conference on Smart Internet of Things (SmartIoT), IEEE, 2020, pp. 176--182.

\bibitem{yuan2022membership}
X.~Yuan, L.~Zhang, Membership inference attacks and defenses in neural network
  pruning, arXiv preprint arXiv:2202.03335 (2022).

\bibitem{molchanov2016pruning}
P.~Molchanov, S.~Tyree, T.~Karras, T.~Aila, J.~Kautz, Pruning convolutional
  neural networks for resource efficient inference, arXiv preprint
  arXiv:1611.06440 (2016).

\end{thebibliography}

\end{document}